\documentclass{article}

\usepackage[main,final]{neurips_2026}

\usepackage[utf8]{inputenc}
\usepackage[T1]{fontenc}
\usepackage{amsmath,amssymb,mathtools}
\usepackage{amsthm}
\usepackage{booktabs}
\usepackage[hidelinks]{hyperref}
\hypersetup{
  colorlinks=false,
  hidelinks,
  citecolor=black,
  linkcolor=black,
  urlcolor=black,
  filecolor=black,
}
\usepackage{cleveref}
\usepackage{xcolor}
\usepackage{enumitem}
\usepackage{bm}
\usepackage{microtype}
\usepackage{tabularx}
\usepackage{array}
\usepackage{graphicx}
\usepackage{wrapfig}
\usepackage{caption}
\usepackage{subcaption}
\usepackage{placeins}
\usepackage{tikz}
\usepackage{afterpage}
\usepackage{titletoc}
\usepackage{needspace}

\titlecontents{section}
  [2.3em]
  {\vspace{0.35em}}
  {\color{blue}\contentslabel{2.3em}}
  {\color{blue}}
  {\color{black}\titlerule*[0.6pc]{.}\contentspage}
\titlecontents{subsection}
  [2.5em]
  {}
  {\color{blue}\contentslabel{2.8em}}
  {\color{blue}}
  {\color{black}\titlerule*[0.6pc]{.}\contentspage}

\newcommand{\R}{\mathbb{R}}
\renewcommand{\v}[1]{\mathbf{#1}}
\newcommand{\param}{\v{\theta}}
\newcommand{\xemb}{\tilde{\v{x}}}

\newcounter{sudokurow}
\newcounter{sudokucol}

\theoremstyle{remark}

\title{One Model, Two Roles: Emergent Specialization in a Shared Recurrent Transformer}

\author{
  Jucheng Shen\thanks{Equal contribution.} \\
  Rice University \\
  \And
  Barbara Su\footnotemark[1] \\
  Stanford University \\
  \And
  Anastasios Kyrillidis\thanks{Corresponding author.} \\
  Rice University \\
}

\begin{document}
\maketitle

\begin{abstract}
Can a shared-weight recurrent Transformer---one that repeatedly applies the same Transformer to update evolving latent states---develop distinct internal roles without being partitioned into separate modules?
We study this in Asymmetric Input Recurrence (AIR), a minimal two-state reasoning architecture in which the same Transformer model is reused for both updates (per literature, L and H) and the only built-in difference in the update rule is that the encoded input is injected during L-updates but not H-updates.
Across Sudoku-Extreme and Maze, decoded rollouts reveal a consistent split: $z_H$ behaves like a fully committed proposal state, whereas $z_L$ retains local uncertainty and shifting intermediate structure.
Freeze interventions provide evidence of dynamical dependence between the two states, with responses that vary across tasks, intervention directions, and seeds.
Ablations show that to induce specialization, the shared model needs to be able to tell the two update types apart, either from input injection asymmetry or from a separate level token.
Mechanistically, attention analysis shows that L-updates are consistently more local than H-updates in both Sudoku and Maze.
Together, these results show that, in a two-state recurrent setting, a clear state-identity signal can induce stable, related functional roles inside a shared-parameter recurrent Transformer.
Code is available at \url{https://github.com/juchengshen/air}.
\end{abstract}

\section{Introduction}\label{sec:intro}

\afterpage{%
\begin{figure}[!t]
    \centering
    \includegraphics[width=\linewidth]{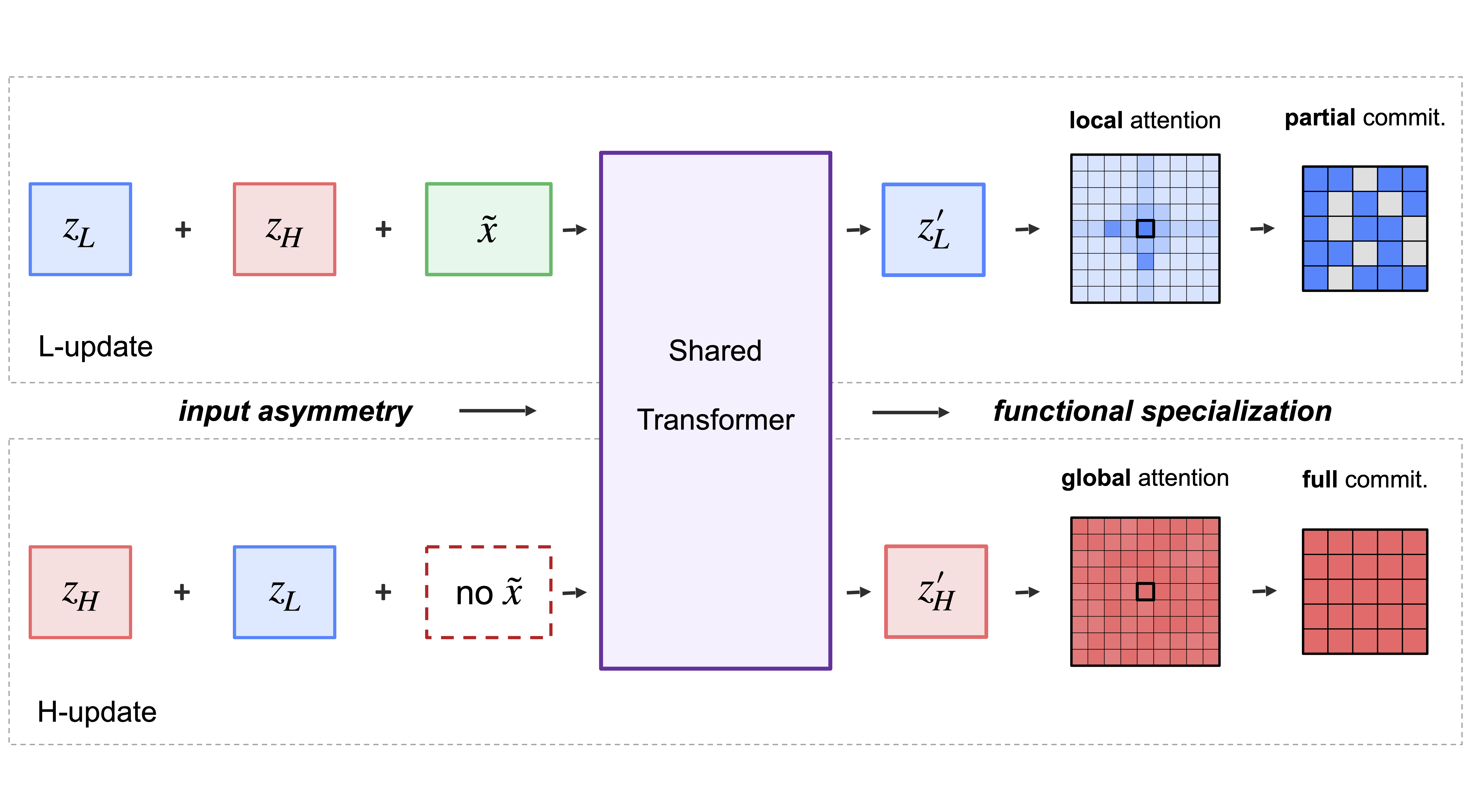}
    \caption{\textbf{AIR testbed.}
    The same shared Transformer $f(\cdot;\param)$ is reused at two positions in the recurrent schedule: with the encoded input $\xemb$ for L-updates (top) and without it for H-updates (bottom).
    Despite identical parameters, this input difference lets the shared model tell the two update types apart: $z_L$ behaves like a partially committed, revision-oriented state while $z_H$ behaves like a committed proposal state, and each query token in L-updates attends to other tokens more locally than H-updates.}
    \label{fig:architecture}
\end{figure}
}

Small recurrent reasoning models are emerging as a promising approach to parameter-efficient reasoning: reusing shared weights increases computational depth without increasing the parameter count.
Concretely, a \emph{recurrent Transformer} reuses the same Transformer block or stack across sequential computation steps to update persistent latent states, so effective computational depth can grow without allocating a separate parameter set at every step; a familiar example is the looped Transformer~\citep{giannou2023looped}, which applies the same block over and over to refine a single evolving latent sequence.
HRM~\citep{wang2025hrm} and TRM~\citep{jolicoeur2025trm} demonstrate strong performance on structured abstract-reasoning benchmarks using two recurrent modules and a single shared-weight recursive network, respectively.
Recent extensions strengthen this direction with stochastic recursive exploration in Probabilistic TRM~\citep{sghaier2026ptrm} and GRAM~\citep{baek2026gram}, and constant-memory fixed-point refinement in Attractor Models~\citep{feinashley2026attractor}, yielding further gains on structured reasoning benchmarks.
Inspired by these models, we raise the following question: \textit{when the same model is reused for two latent states, does it learn distinct computational roles or merely process both states in the same way?}
The answer matters because it tests whether specialization needs separate parameter models, or whether one shared computation can specialize when it gets a stable signal telling it which state it is updating.

To study this question, we introduce  Asymmetric Input Recurrence (AIR), a minimal testbed that keeps TRM's shared recurrent model but uses HRM's one-step truncated gradient.
We use this approximation for memory efficiency: backpropagating only through the final state updates avoids storing the full recurrent trajectory and keeps activation-memory cost approximately constant with recurrence depth~\citep{wang2025hrm,jolicoeur2025trm}.
Stronger recurrent reasoners share this systems choice: GRAM~\citep{baek2026gram} keeps HRM's truncated propagation and Attractor Models~\citep{feinashley2026attractor} use constant-memory implicit gradients, so AIR's diagnostics yield hypotheses testable in these systems (\Cref{sec:conclusion}).
In AIR, the only built-in difference in the update rule is whether the encoded input $\xemb$ is injected (\Cref{fig:architecture}), so AIR is a controlled setting for testing whether this single difference in update content can induce a division of labor between a high-level latent state $z_H$ and a low-level latent state $z_L$.

Long story short, our answer is yes for the cases this paper studies.
Across Sudoku-Extreme~\citep{sudokuextreme2025} and Maze~\citep{mazehard2025}, decoded rollouts show a stable split in which $z_H$ behaves like a committed proposal state and $z_L$ retains local uncertainty and shifting intermediate structure.
We call $z_H$ \emph{fully committed} since it always decodes to a complete candidate, and $z_L$ \emph{partially committed} since some of its positions stay undecided.

Freeze experiments, used as an out-of-distribution stress test, show that the two states are dynamically interdependent (\Cref{sec:causality}), and attention analysis reveals a matching mechanism: L-updates are more local and, on Sudoku, more violation-sensitive, whereas H-updates are more global (\Cref{sec:mechanism}).

Our key contributions are as follows:
\begin{enumerate}[leftmargin=*]
    \item \textbf{Existence of functional specialization.} We demonstrate the existence of functional specialization in a shared-parameter recurrent Transformer. When decoding the intermediate output of each state, we observe that AIR develops a stable committed-$z_H$ / partially uncommitted-$z_L$ split, with a matching L-local versus H-global attention pattern (\Cref{sec:existence,sec:mechanism}).
    \item \textbf{State-identity signal can induce functional specialization.} We empirically show that, under the training setting studied, functional specialization emerges when the shared model can tell the two update types apart---either through input injection asymmetry or through a separate level token (\Cref{sec:injection}). Input asymmetry consistently improves Sudoku accuracy, while Maze shows a positive average trend.
    \item \textbf{Two states are related.} Freeze interventions provide evidence of dynamical dependence between the two states, with responses that vary across tasks, intervention directions, and seeds (\Cref{sec:causality}).
\end{enumerate}

\section{Related work}\label{sec:related}

\textbf{Iterative and recurrent reasoning.}
AIR belongs to work that uses recurrence or weight-sharing to turn small networks into multi-step reasoners.
Universal Transformers~\citep{dehghani2019universal} and later iterated models showed that weight sharing can support flexible computation and adaptive depth~\citep{giannou2023looped,schwarzschild2021learn,banino2021pondernet}.
Its closest predecessors are HRM~\citep{wang2025hrm} and TRM~\citep{jolicoeur2025trm}: HRM assigns the two latent states to different parameter models, whereas TRM ties them to one shared model.
AIR keeps the shared-model setting but asks whether distinct roles can still emerge inside that shared recurrence when input reaches at least one state, and when the model can tell which state it is updating.
\citet{blayney2026mechanistic} show that input re-injection can prevent degeneration in looped language models; we ask whether \emph{asymmetric} input injection can also help a shared model separate two recurrent roles.

\textbf{Modularity and emergent specialization.}
Our question is also about the distinction between \emph{structural modularity} and \emph{functional specialization}.
Neural Module Networks~\citep{andreas2016neural} and Recurrent Independent Mechanisms~\citep{goyal2021recurrent} impose modular structure by design, whereas work on emergent modularity studies when differentiated behavior can arise inside initially monolithic models~\citep{csordas2021neural,tomoda2025emergence}.
AIR isolates this issue in a particularly strict setting: because the recurrent model is shared, any division of labor must emerge from training rather than architecture.
Conditioning a shared network on a mode signal is also well established, e.g., conditional and unconditional roles in classifier-free guidance~\citep{ho2022classifier} and task or target-language tokens in multitask and multilingual models~\citep{johnson2017google,raffel2020exploring}; 
there, however, each mode is defined and trained externally, whereas AIR trains on a single task with every training signal read from $z_H$: the update types can be told apart (by input injection or a level token), but nothing specifies what each state should compute.
Thus our novelty is neither shared recurrence nor conditional computation by itself, but the controlled demonstration, ablation, and mechanistic characterization of a role split that emerges without any supervision of what each recurrent state should do.

\textbf{Mechanistic analysis and latent workspaces.}
Methodologically, our paper is closest to mechanistic and perturbation-based analyses of reasoning models~\citep{elhage2021mathematical,olsson2022context,brinkmann2024mechanistic,hong2024transformers}, but we analyze recurrent latent states rather than individual heads or circuits.
It also connects to work on internal workspaces for reasoning: unlike token-level scratchpads or chain-of-thought~\citep{nye2021show,wei2022chain}, AIR resembles latent-reasoning approaches such as COCONUT~\citep{hao2024coconut}, with $z_L$ behaving like a partial workspace and $z_H$ like a more committed proposal state.
The difference is that AIR does not prescribe these roles; it asks whether they emerge in a shared-weight recurrent Transformer when the model can tell the two states apart.

\section{Preliminaries}\label{sec:preliminaries}

We compare three closely related recurrent architectures: HRM, TRM, and our variant AIR.
All maintain two latent states, $z_H, z_L \in \R^{B \times S \times D}$, where $B$ is batch size, $S$ is sequence length, and $D$ is hidden dimension.
Let $\xemb$ denote the encoded input.

HRM~\citep{wang2025hrm}, TRM~\citep{jolicoeur2025trm}, and AIR all use the following two-state recurrence in general; their specific differences are described below:
\begin{align}
    \text{\textbf{L-update:}} \quad z_L &\leftarrow f_L\!\left(z_L + z_H + n_L \xemb;\,\param_L\right), \label{eq:hrm_l}\\
    \text{\textbf{H-update:}} \quad z_H &\leftarrow f_H\!\left(z_H + z_L + n_H \xemb;\,\param_H\right). \label{eq:hrm_h}
\end{align}
Here $n_L$ and $n_H$ are injection counts: the number of copies of $\xemb$ added in each L- and H-update.
Unless otherwise stated, all three variants use $(n_L,n_H)=(1,0)$, so the L-update receives $\xemb$ and the H-update receives no direct input.
Each high-level cycle interleaves $C_L$ L-updates with one H-update and repeats this pattern $C_H$ times.

The first difference between HRM, TRM, and AIR is whether the L- and H-updates use one or two networks.
HRM uses separate Transformer models, $f_L$ and $f_H$, with parameter sets $\param_L$ and $\param_H$.
TRM and AIR tie the two updates into a single shared model, $f_L=f_H=f$.
Like looped Transformers, all three models reuse the same computation at every recurrence step (the same category in the taxonomy of \citet{mozer2026topological}); the difference is that a looped Transformer refines a single latent state, whereas these models maintain two, $z_L$ and $z_H$, and each update takes both as input (\Cref{eq:hrm_l,eq:hrm_h}).

The second difference is how the recurrence is trained.
HRM and AIR use a one-step truncated gradient~\citep{williams1990efficient}: gradients are computed through only one L-update and one H-update per training step.
TRM trains by backpropagating through one full final recursion.
HRM also uses ACT-style halting~\citep{graves2016adaptive}, but following the observation of \citet{ge2025hierarchicalreasoningmodelsperspectives} that evaluation usually runs to the maximum cycle count, we do not analyze ACT here.

\textbf{Notations.}
Unless otherwise stated, we decode intermediate states with the same output head used at test time.
Throughout the rest of the paper, we refer to each individual L- and H-update as a \emph{sub-step}.
Following HRM, we use the default $C_L = C_H = 2$, giving the sub-step pattern \textbf{L, L, H, L, L, H} within each cycle.
The first four sub-steps run without gradient; gradient is computed only through the final L-update and the final H-update.
All experiments use Sudoku-Extreme ($9 \times 9$ puzzles with 17 givens) and Maze ($30 \times 30$ grids); full implementation and training details are in \Cref{app:experimental_details}.

\section{Emergent functional specialization}\label{sec:existence}

This section asks whether AIR's minimal asymmetry---input injection in L-updates but not H-updates---can produce distinct computational roles.
We test this directly by decoding both latent states, $z_H$ and $z_L$, at every sub-step with the same output head used at test time.

\begin{figure}[!htbp]
    \centering
    \includegraphics[width=\textwidth]{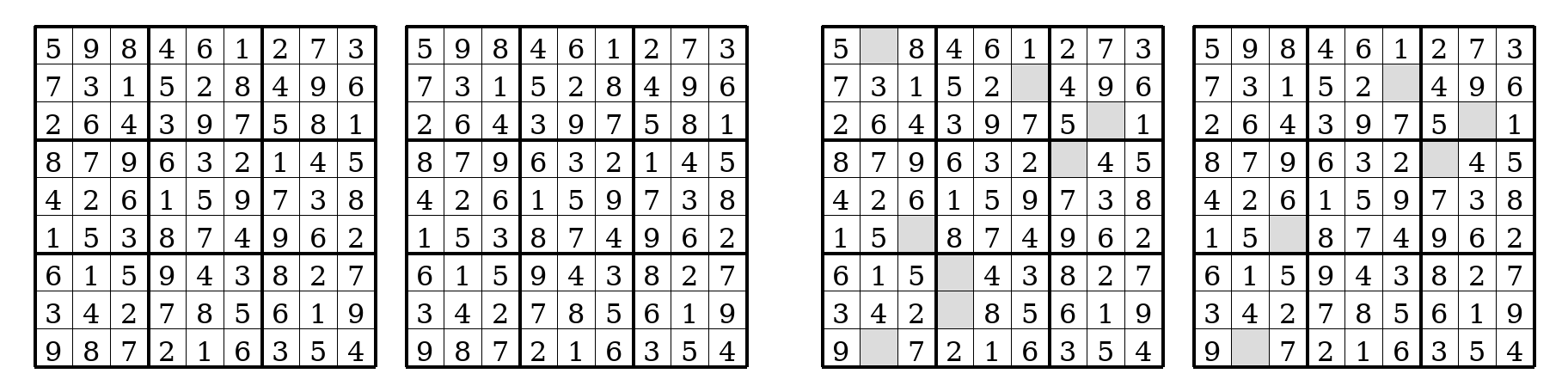}
    \caption{\textbf{Sudoku decoded states.}
    Left = $z_H$, Right = $z_L$;
    $z_H$ stays fully committed, while $z_L$ keeps some cells as \texttt{BLANK} (shaded \textcolor[HTML]{B0B0B0}{\textbf{light gray}}) and shifts those blanks across sub-steps.}
    \label{fig:zH_zL_sudoku}
\end{figure}

\begin{figure}[!htbp]
    \centering
    \includegraphics[width=\textwidth]{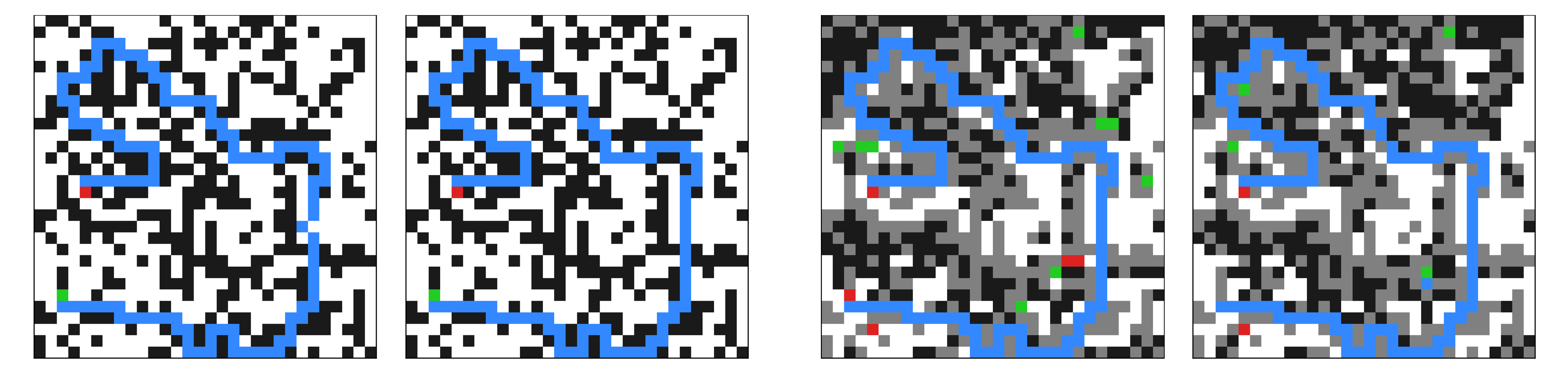}
    \caption{\textbf{Maze decoded states.}
    Left = $z_H$, Right = $z_L$;
    \textcolor[RGB]{26,26,26}{\textbf{Black}} = wall, white = open path, \textcolor[RGB]{51,136,255}{\textbf{blue}} = solution path, \textcolor[RGB]{128,128,128}{\textbf{gray}} = undecided (\texttt{PAD}), \textcolor[RGB]{221,34,34}{\textbf{red}} = goal, and \textcolor[RGB]{34,204,34}{\textbf{green}} = start.
    $z_H$ stays committed, while $z_L$ shows \textcolor[RGB]{128,128,128}{\textbf{gray}} undecided cells and other shifting local structure. Multiple cells are decoded as start or goal in the $z_L$ panels, which may reflect endpoint information that is not yet stably localized in these intermediate states.}
    \label{fig:zH_zL_maze}
\end{figure}

The asymmetric AIR rollouts reveal the same qualitative split on both tasks.
$z_H$ is \emph{fully committed}: at every sub-step it decodes to a complete candidate solution, even when that candidate is still wrong.
$z_L$, by contrast, is \emph{partially committed}: some positions remain undecided, and the locations of those undecided positions move across sub-steps.
For Sudoku, the undecided symbol is the dedicated \texttt{BLANK} token; for Maze, it is the analogous \texttt{PAD} token.
Both commitment terms describe only how the decoded states look: since only $z_H$ is directly supervised and $z_L$ is read through the same output head, they do not by themselves establish what computation each state performs.
\Cref{fig:zH_zL_sudoku,fig:zH_zL_maze} visualize this split.
These panels show a representative rollout from one checkpoint per task; inspecting decoded rollouts from all five independently trained checkpoints per task, we observe the same qualitative split in every seed.

\begin{figure}[!t]
    \centering
    \includegraphics[width=\textwidth]{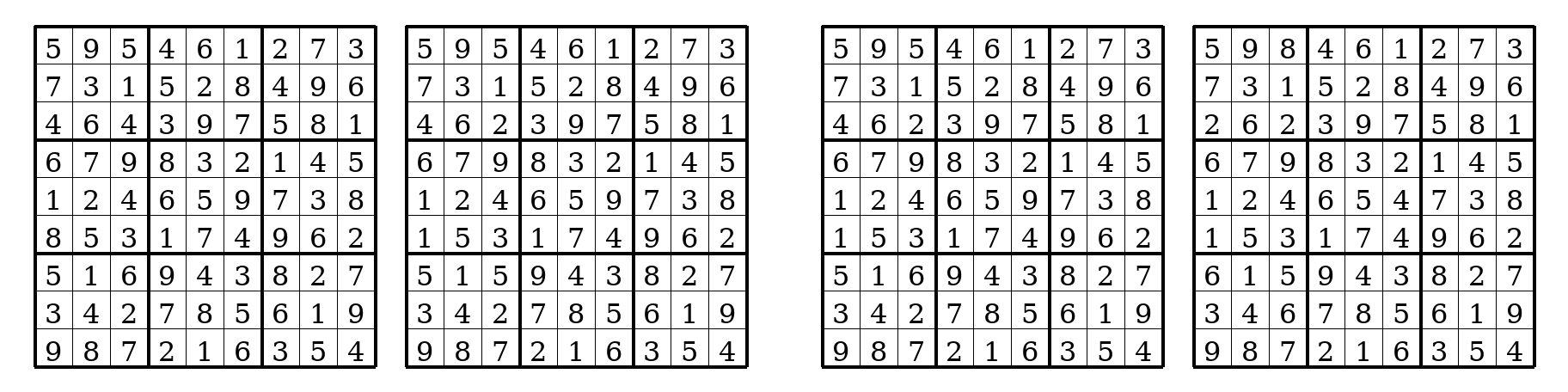}
    \caption{\textbf{Sudoku decoded states for the symmetric \texttt{Lx\_Hx} variant.}
    Left = $z_H$, Right = $z_L$;
    decoded intermediate states across two sub-steps for each update type. Relative to \Cref{fig:zH_zL_sudoku}, decoded $z_L$ no longer shows visible \texttt{BLANK} cells.}
    \label{fig:zH_zL_sudoku_symmetric}

    \vspace{6pt}
    \includegraphics[width=\textwidth]{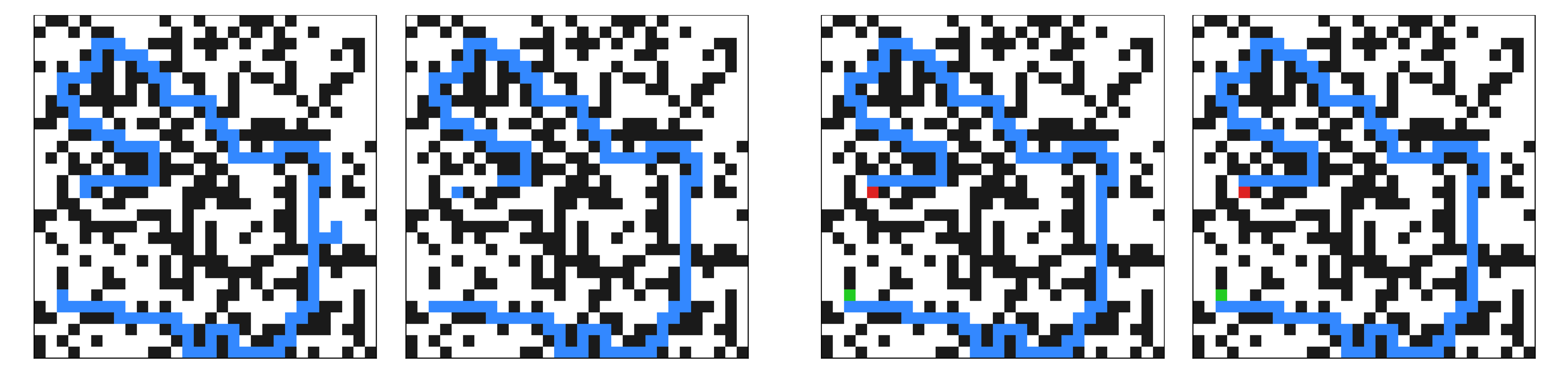}
    \caption{\textbf{Maze decoded states for the symmetric \texttt{Lx\_Hx} variant.}
    Left = $z_H$, Right = $z_L$;
    decoded intermediate states across two sub-steps for each update type. Relative to \Cref{fig:zH_zL_maze}, decoded $z_L$ no longer shows visible \texttt{PAD} / undecided regions.}
    \label{fig:zH_zL_maze_symmetric}
\end{figure}
\FloatBarrier

The symmetric \texttt{Lx\_Hx} control removes this decoded role split.
\Cref{fig:zH_zL_sudoku_symmetric,fig:zH_zL_maze_symmetric} repeat the same four-panel view for symmetric models on Sudoku and Maze.
Once the asymmetric input difference is removed, decoded $z_L$ no longer shows the visible \texttt{BLANK} / \texttt{PAD} undecided regions that characterize AIR's partially committed state, and decoded $z_H$ no longer reaches the same degree of early commitment seen in the asymmetric AIR rollouts.

Together, the asymmetric and symmetric rollouts show that functional specialization is not simply a consequence of carrying two recurrent states.
The same shared Transformer develops a proposal-like $z_H$ and a locally uncertain $z_L$ when the update types are distinguishable through input injection, but this decoded division nearly vanishes when both update types receive the same input treatment.
This comparison motivates the ablations and mechanisms studied next: specialization emerges when the shared model has a reliable state-identity signal for the two latent update types.

\section{Input injection and state identity}\label{sec:injection}

\Cref{sec:existence} showed that AIR develops distinct $z_H$/$z_L$ roles.
This section asks what signal allows a single shared model to maintain that split.
Our hypothesis is simple: the shared model must have a way to tell which state it is updating. Since the only visible difference between the two update types are whether input $\xemb$ is injected, it is reasonable to test how many copies of $\xemb$ are added to each update type.
We parameterize each configuration by the injection counts $(n_L,n_H)$ in \Cref{eq:hrm_l,eq:hrm_h}, and define the asymmetry level as $\Delta = |n_L - n_H|$.

\begin{table}[t]
\centering
\caption{Injection asymmetry ablation on Sudoku-Extreme and Maze. Exact-solve accuracy, mean $\pm$ SD over five independently trained seeds per shared-model variant.
}
\label{tab:injection_sudoku}
\label{tab:injection_maze}
\begin{tabular}{llccc}
\toprule
\textbf{Variant} & $(n_L, n_H)$ & $\Delta$ & \textbf{Sudoku (\%)} & \textbf{Maze (\%)} \\
\midrule
\multicolumn{5}{l}{\emph{Asymmetric ($\Delta > 0$)}} \\[2pt]
L\_Hx               & $(0,\,1)$ & 1 & $58.7_{\pm 3.3}$ & $75.3_{\pm 3.2}$ \\
Lx\_H (default)     & $(1,\,0)$ & 1 & \textbf{$60.0_{\pm 2.0}$} & $71.0_{\pm 6.3}$ \\
L\_H2x              & $(0,\,2)$ & 2 & $58.6_{\pm 1.9}$ & \textbf{$75.6_{\pm 1.9}$} \\
L2x\_H              & $(2,\,0)$ & 2 & $59.1_{\pm 2.4}$ & $71.1_{\pm 6.4}$ \\
Lx\_H2x             & $(1,\,2)$ & 1 & $59.6_{\pm 0.9}$ & $70.9_{\pm 2.4}$ \\
L2x\_Hx             & $(2,\,1)$ & 1 & $58.6_{\pm 2.9}$ & $74.5_{\pm 1.6}$ \\
\addlinespace[2pt]
\multicolumn{3}{l}{\emph{Approx. group mean}} & $59.1$ & $73.1$ \\
\midrule
\multicolumn{5}{l}{\emph{Symmetric ($\Delta = 0$)}} \\[2pt]
Lx\_Hx              & $(1,\,1)$ & 0 & $52.1_{\pm 1.6}$ & $69.4_{\pm 2.5}$ \\
L2x\_H2x            & $(2,\,2)$ & 0 & $50.9_{\pm 2.9}$ & $70.3_{\pm 4.2}$ \\
\addlinespace[2pt]
\multicolumn{3}{l}{\emph{Approx. group mean}} & $51.5$ & $69.9$ \\
\midrule
\multicolumn{5}{l}{\emph{Two-network baseline (separate parameters)}} \\[2pt]
HRM~\citep{wang2025hrm} & --- & --- & $55.0$ & $74.5$ \\
\bottomrule
\end{tabular}
\end{table}

\textbf{Asymmetric input injection can induce functional specialization.}
\mbox{\Cref{tab:injection_sudoku}} shows a consistent accuracy advantage on Sudoku and a positive average trend on Maze.
On Sudoku, asymmetric configurations ($\Delta > 0$) cluster around $59\%$, whereas symmetric ones ($\Delta = 0$) fall to about $51\%$, an ${\sim}8$-point drop.
All 12 asymmetric-versus-symmetric Sudoku comparisons remain significant after Holm correction.
On Maze, the descriptive group means are $73.1\%$ versus $69.9\%$, but only one of 12 comparisons survives correction (L\_H2x versus Lx\_Hx, adjusted $p=0.0318$).
One thing to note is that the best AIR variants are also comparable to---and on Sudoku even exceed---the original two-network HRM baseline ($55.0\%$ on Sudoku, $74.5\%$ on Maze) with only 50\% of the latter's parameters, demonstrating the promise of this ``one model, two roles'' paradigm when designing efficient future LLM architectures.
Detailed significance tests for Table~\ref{tab:injection_maze} can be seen in Appendix Section \mbox{\ref{app:table1_significance}}.

\textbf{Level-token control.}
A natural alternative explanation is that AIR does not need asymmetric \emph{input} specifically; perhaps input can be injected symmetrically as long as a separate signal tells the model which state it is updating.
To test this, we start from the symmetric Lx\_Hx baseline ($(1,1)$, $\Delta = 0$), which preserves L's single input copy and gives H the same input, and add learned level tokens in three ways:
\begin{enumerate}[label=(\roman*), leftmargin=*]
    \item \textbf{Addition.} A learned vector $\tau_L$ or $\tau_H$ is added element-wise to every token before each sub-step.
    \item \textbf{Prepend (strip).} A learned level token is prepended at position~0, participates in self-attention during the forward pass, and is then stripped from the output after the final Transformer layer. In the next sub-step, the same level token is prepended again and repeat.
    \item \textbf{Prepend (no strip).} The same prepended token is not stripped after the final Transformer layer; it stays at the beginning of the sequence across cycles.
\end{enumerate}

\begin{table}[!htbp]
\centering
\caption{Level-token controls on Sudoku-Extreme with symmetric Lx\_Hx injection. Mean $\pm$ sample SD over five seeds, using each run's best evaluation within 20,000 epochs.}
\label{tab:level_tokens}
\begin{tabular}{lcc}
\toprule
\textbf{Level-token strategy} & \textbf{Mechanism} & \textbf{Acc.\ (\%)} \\
\midrule
Lx\_Hx (base, no token)     & ---                 & $52.07_{\pm 1.62}$ \\
\quad + Addition              & element-wise add    & $50.91_{\pm 3.02}$ \\
\quad + Prepend (strip)       & prepend, attend, strip & $58.23_{\pm 1.56}$ \\
\quad + Prepend (no strip)    & prepend, persist       & $55.34_{\pm 5.52}$ \\
\midrule
\multicolumn{2}{l}{\emph{For reference:} asymmetric $\Delta = 1$ variants} & $\sim\!59.0$ \\
\bottomrule
\end{tabular}
\end{table}

In \mbox{\Cref{tab:level_tokens}}, prepend (strip) raises exact solve from $52.07\%$ to $58.23\%$.
We compare each seed's best evaluation over ten evenly-spaced checkpoints using two-sided Welch tests, with Holm correction across all six pairwise comparisons, and report 95\% confidence intervals for each difference in accuracy points.
Strip significantly outperforms Base and Addition ($p_{\mathrm{adj}}=0.00168$ and $0.0149$), while Addition does not differ significantly from Base ($p_{\mathrm{adj}}=0.788$).
No-strip reaches $55.34\pm5.52\%$ but does not differ significantly from any other control ($p_{\mathrm{adj}}=0.788$, $0.659$, and $0.788$ against Base, Addition, and Strip); all six differences and their 95\% confidence intervals are in \Cref{tab:level_token_tests}.

Thus, a transient, structurally separable level token supports recovery under symmetric input injection, whereas Addition shows no detected improvement.
Addition mixes identity into every content token; no-strip allows its dedicated slot to accumulate recurrent content, but its higher variance leaves the comparison with Strip inconclusive.
A more detailed interpretation of why addition fails while prepend (strip) works is deferred to \Cref{app:level_token_interpretation}.
Longer-training curves and token-consistency analysis are provided in Appendix Section \mbox{\ref{app:level_token_training}}.

\textbf{Operator-form control.}
We also test linear, nonlinear, sign-flip, and Hadamard transformations of $\xemb$ before injection; on Sudoku, none differs significantly from additive injection after Holm correction across four comparisons (details in \mbox{\Cref{app:operator_forms}}).

\section{State-dependence analysis}\label{sec:causality}

In this section, we analyze state dependence: we test whether the two states in AIR depend on each other dynamically by applying paired freeze experiments: hold $z_H$ fixed while $z_L$ updates, or hold $z_L$ fixed while $z_H$ updates, and measure how the unfrozen state changes.

\textbf{Metric: content change.}
We quantify state dependence by \emph{content change}, the number of positions whose decoded token changes between consecutive sub-steps.
This measures how much one state keeps revising its decoded proposal when the other state is prevented from updating.
Writing $\hat{y}_i^{(t)}$ for the decoded token at position $i$ after sub-step $t$, we define content change $C^{(t)}$ as
\[
  C^{(t)} = \bigl|\bigl\{i : \hat{y}_i^{(t)} \neq \hat{y}_i^{(t-1)}\bigr\}\bigr|.
\]
All content-change plots report mean $\pm$ sample standard deviation across three seeds, with 1{,}000 evaluated puzzles per seed; per-seed results are in \Cref{app:per_seed_freeze}.

\begin{figure}[t]
    \centering
    \begin{subfigure}[b]{0.48\textwidth}
        \centering
        \includegraphics[width=\textwidth]{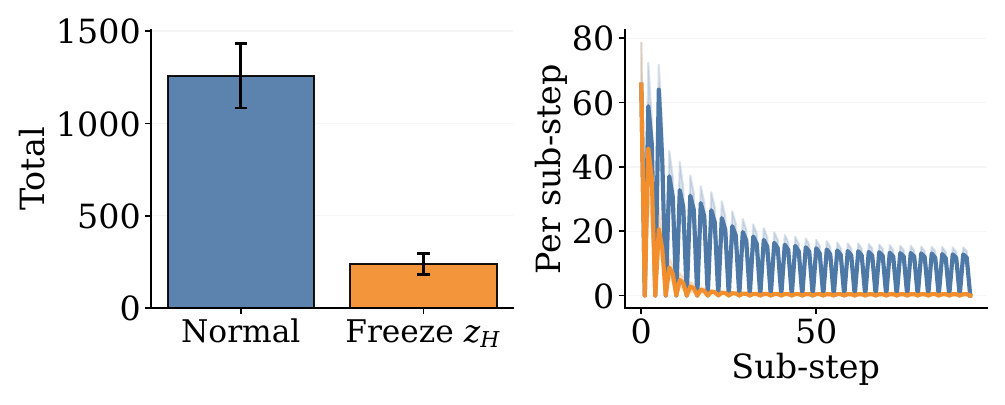}
        \caption{Sudoku: freezing $z_H$ reduces $z_L$'s changes.}
        \label{fig:freeze_zH_sudoku}
    \end{subfigure}
    \hfill
    \begin{subfigure}[b]{0.48\textwidth}
        \centering
        \includegraphics[width=\textwidth]{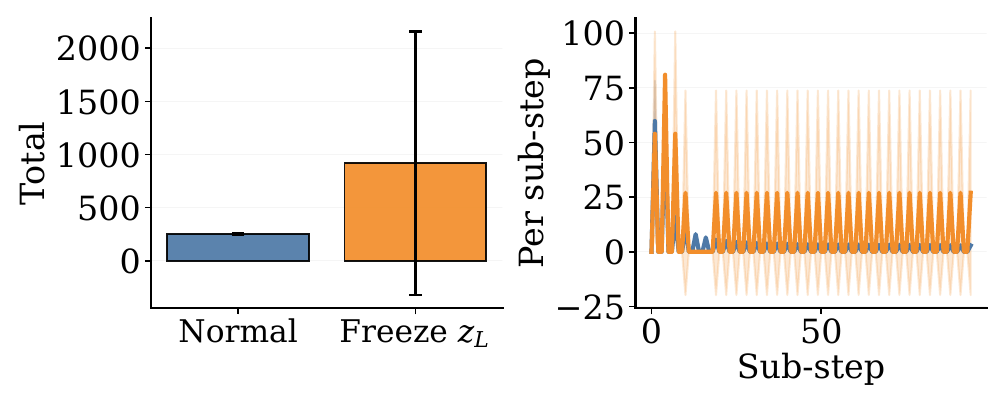}
        \caption{Sudoku: freezing \(z_L\) has a seed-dependent effect on \(z_H\)’s changes.}
        \label{fig:freeze_zL_sudoku}
    \end{subfigure}

    \vspace{0.5em}

    \begin{subfigure}[b]{0.48\textwidth}
        \centering
        \includegraphics[width=\textwidth]{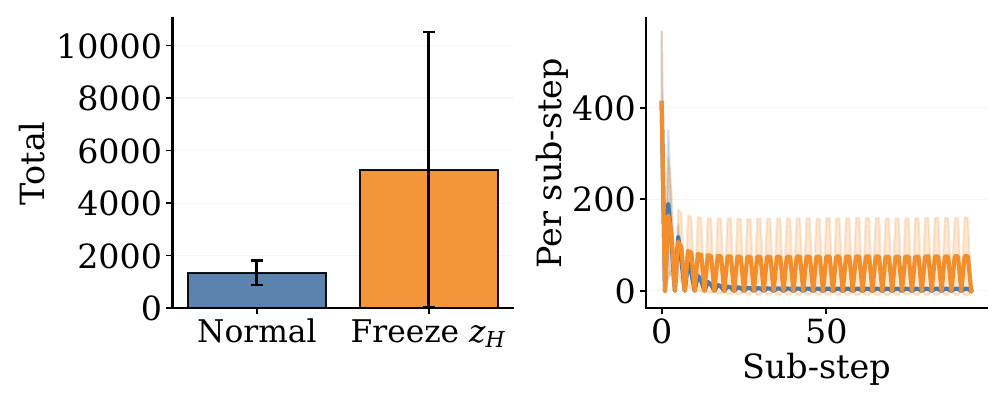}
        \caption{Maze: freezing $z_H$ increases $z_L$'s changes, with substantial variation across seeds.}
        \label{fig:freeze_zH_maze}
    \end{subfigure}
    \hfill
    \begin{subfigure}[b]{0.48\textwidth}
        \centering
        \includegraphics[width=\textwidth]{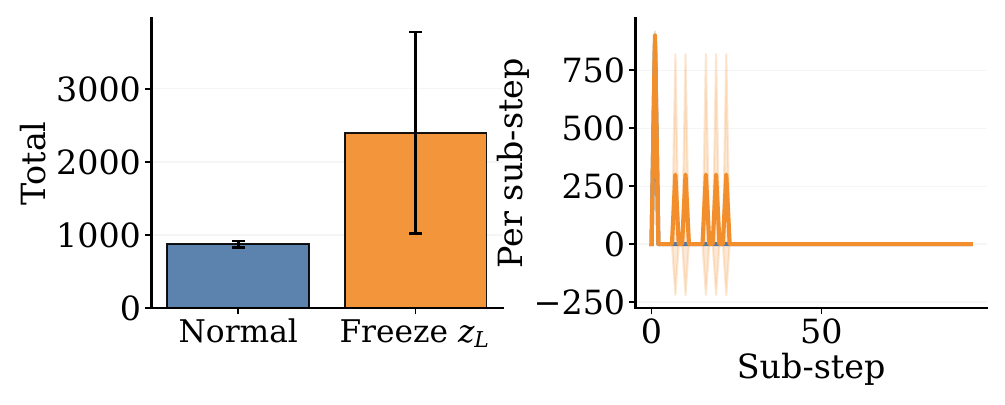}
        \caption{Maze: freezing $z_L$ increases $z_H$'s changes.}
        \label{fig:freeze_zL_maze}
    \end{subfigure}
    \caption{Paired freeze experiments on AIR (asymmetric \texttt{Lx\_H}) measured by content changes. Bars show total content changes per puzzle and curves show per-sub-step changes, both as mean $\pm$ sample SD across three independently trained seeds (1{,}000 evaluated puzzles per seed).}
    \label{fig:freeze_interventions}
\end{figure}

\textbf{Sudoku: freezing the two states has different effects.}
In Sudoku, freezing either latent state's update disrupts the other's normal dynamics, with responses that depend on the frozen state and the seed.
Freezing $z_H$ sharply suppresses $z_L$'s activity, reducing total content changes from $1258.2$ to $239.5$ and lowering the per-sub-step change after the first few steps (\Cref{fig:freeze_zH_sudoku}).
Freezing $z_L$ produces heterogeneous responses in $z_H$: total content changes increase on average from $250.1$ to $918.0$, but this increase is driven by one of three seeds, while the other two show decreases (\Cref{fig:freeze_zL_sudoku}).

\textbf{Maze: freezing one state increases content changes in the other.}
Maze shows the same state interdependence with the opposite pattern: freezing one latent state's update increases the other's content changes.
Freezing $z_H$ increases $z_L$'s total content changes in all three evaluated seeds, from $1345.8$ to $5277.1$ on average, although the magnitude varies substantially and the mean is strongly influenced by one seed (\Cref{fig:freeze_zH_maze}).
Freezing $z_L$ increases $z_H$'s content changes from $871.9$ to $2400.0$ (\Cref{fig:freeze_zL_maze}).

\begin{figure}[t]
    \centering
    \begin{subfigure}[b]{0.48\textwidth}
        \centering
        \includegraphics[width=\linewidth]{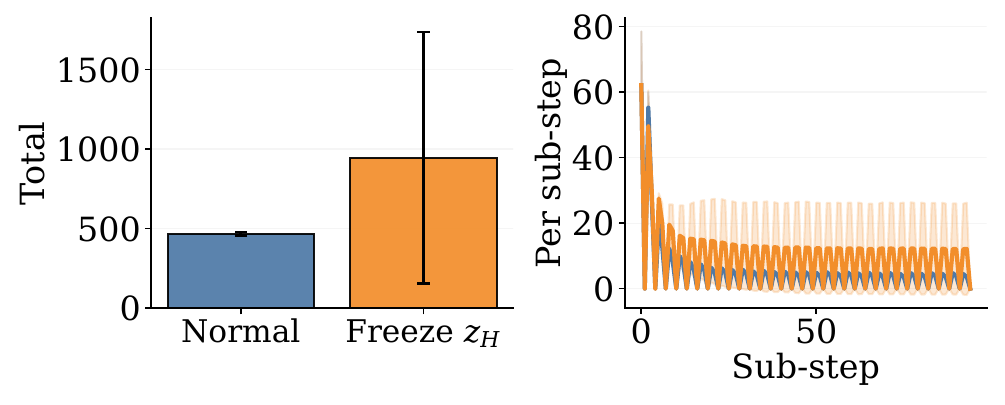}
        \caption{Sudoku: freeze $z_H$ in the symmetric \texttt{Lx\_Hx} model.}
    \end{subfigure}
    \hfill
    \begin{subfigure}[b]{0.48\textwidth}
        \centering
        \includegraphics[width=\linewidth]{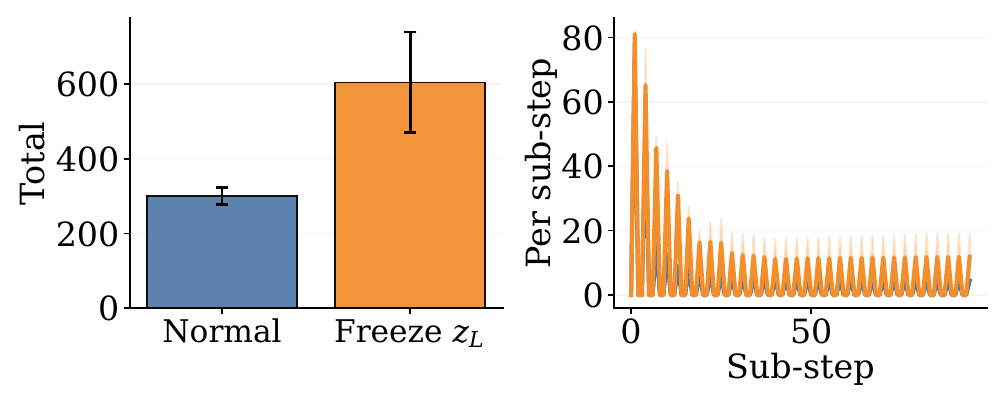}
        \caption{Sudoku: freeze $z_L$ in the symmetric \texttt{Lx\_Hx} model.}
    \end{subfigure}

    \vspace{0.5em}

    \begin{subfigure}[b]{0.48\textwidth}
        \centering
        \includegraphics[width=\linewidth]{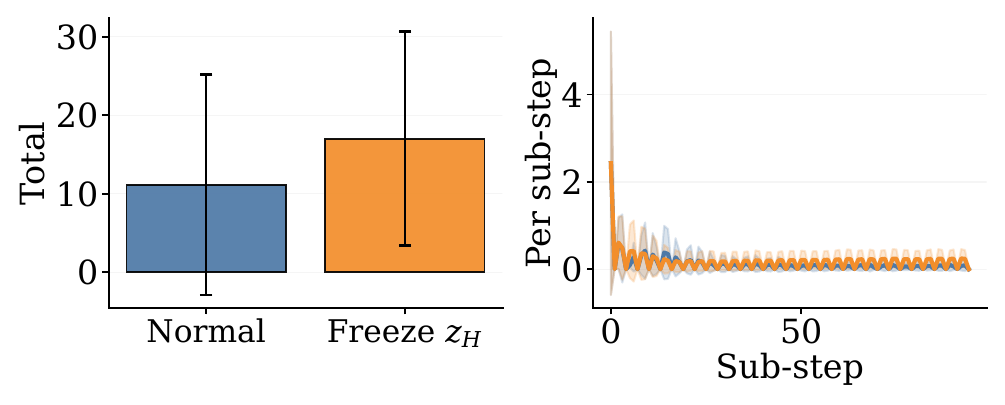}
        \caption{Maze: freeze $z_H$ in the symmetric \texttt{Lx\_Hx} model.}
    \end{subfigure}
    \hfill
    \begin{subfigure}[b]{0.48\textwidth}
        \centering
        \includegraphics[width=\linewidth]{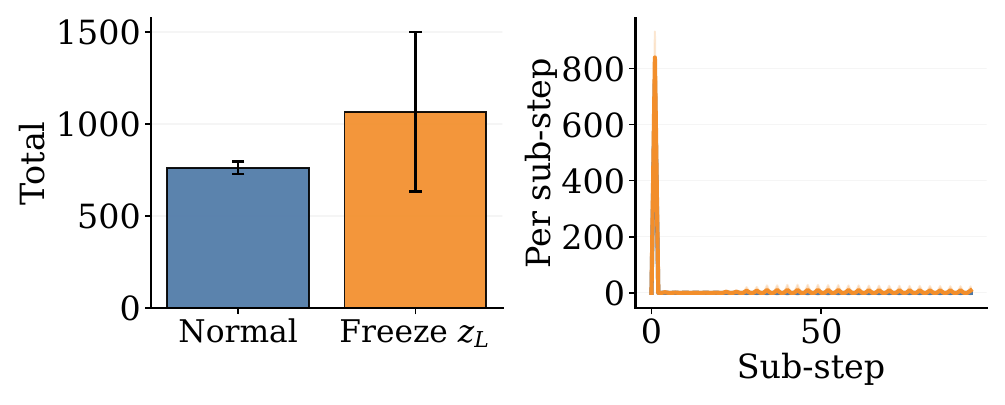}
        \caption{Maze: freeze $z_L$ in the symmetric \texttt{Lx\_Hx} model.}
    \end{subfigure}
    \caption{Freeze experiments on the symmetric \texttt{Lx\_Hx} control on Sudoku and Maze, measured by content changes as mean $\pm$ sample SD across three independently trained seeds (1{,}000 evaluated puzzles per seed).}
    \label{fig:symmetric_freeze_main}
\end{figure}

The symmetric \texttt{Lx\_Hx} control responds differently (\Cref{fig:symmetric_freeze_main}).
On Sudoku, freezing either state increases content changes relative to normal on average.
On Maze, the two directions are qualitatively mismatched: freeze-$z_H$ produces near-zero content-change activity, while freeze-$z_L$ produces a large first-step spike.
Thus the symmetric states are also perturbation-sensitive but do not reproduce AIR's task-dependent freeze-$z_H$ pattern.

\textbf{Effect on task accuracy.}
Freezing either $z_H$ or $z_L$ also collapses final accuracy to $0\%$ on both tasks; on the same 1{,}000-puzzle evaluation subset, the unfrozen setting reaches $55.1\%$ on Sudoku and $71.0\%$ on Maze.

\section{Mechanism: local versus global attention}
\label{sec:mechanism}

\Cref{sec:existence} showed that AIR develops distinct latent-state roles, and \Cref{sec:causality} showed that the two states are dynamically interdependent.
Here we ask whether this split appears in attention.
We test whether L-updates attend more locally than H-updates, and whether L-updates place extra attention on currently violated cells.
On Sudoku, both effects appear: the local/global split is present at every layer, while violation-specific attention appears mainly in deeper layers.
On Maze, we find the local/global split but not a separate violation-specific signal.

\subsection{Setup}\label{sec:mechanism_setup}

\paragraph{Cells, neighborhoods, and violations.}
A \emph{query cell} $q$ is any blank cell.
For Sudoku, its \emph{neighborhood} $\mathcal{N}(q)$ is the 20 cells sharing a row, column, or $3{\times}3$ box with $q$.
A cell is \emph{violated} if its currently decoded value from $z_H$ clashes with another cell in its row, column, or box; let $\mathcal{V}$ denote the set of such cells.
We call a query \emph{violation-adjacent} if $\mathcal{N}(q)\cap\mathcal{V}\neq\emptyset$, and \emph{control} otherwise.
For Maze, we use the analogous setup on the $30{\times}30$ grid.
We define $\mathcal{N}_4(q)$ as the four cardinal neighbors of $q$, $\mathcal{N}_8(q)$ as the eight surrounding neighbors, and $\mathcal{N}_{5\times5}(q)$ as the centered $5{\times}5$ window around $q$ excluding $q$ itself.
Since these neighborhoods have different sizes, we use the per-cell density contrast $\Delta\rho^{(k)}$, which measures the L$-$H attention-mass difference inside $\mathcal{N}_k(q)$ divided by $|\mathcal{N}_k(q)|$.
Maze queries are \emph{error-adjacent} if $\mathcal{N}_4(q)$ contains a cell where decoded $z_H$ disagrees with the ground truth, and \emph{control} otherwise. Details are in \Cref{app:attention_viz,app:maze_attention} for Sudoku and Maze respectively.

\paragraph{What we measure.}
We compare attention during the first L-update and first H-update at cycles $\{2,4,6,8,10,12,14,15\}$, averaging over the 8 heads at each layer.
For Sudoku, we report three L$-$H contrasts:
$\Delta_{\mathrm{nbr}}$, the difference in attention mass on the neighborhood (nbr) $\mathcal{N}(q)$;
$\Delta_{\mathrm{ent}}$, the H$-$L difference in attention entropy, so positive values mean L is more concentrated;
and $\Delta_{\mathrm{viol}}$, the difference in attention mass on $\mathcal{N}(q)\cap\mathcal{V}$.
For Maze, we report $\Delta\rho^{(k)}$, $\Delta_{\mathrm{ent}}$, and $\Delta_{\mathrm{viol}}$ analogously.
Full per-layer values are in \Cref{app:attention_layer_robustness} for Sudoku and \Cref{sec:maze_layers} for Maze.

\subsection{Results}

\begin{figure}[!htp]
    \centering
    \includegraphics[width=0.92\linewidth]{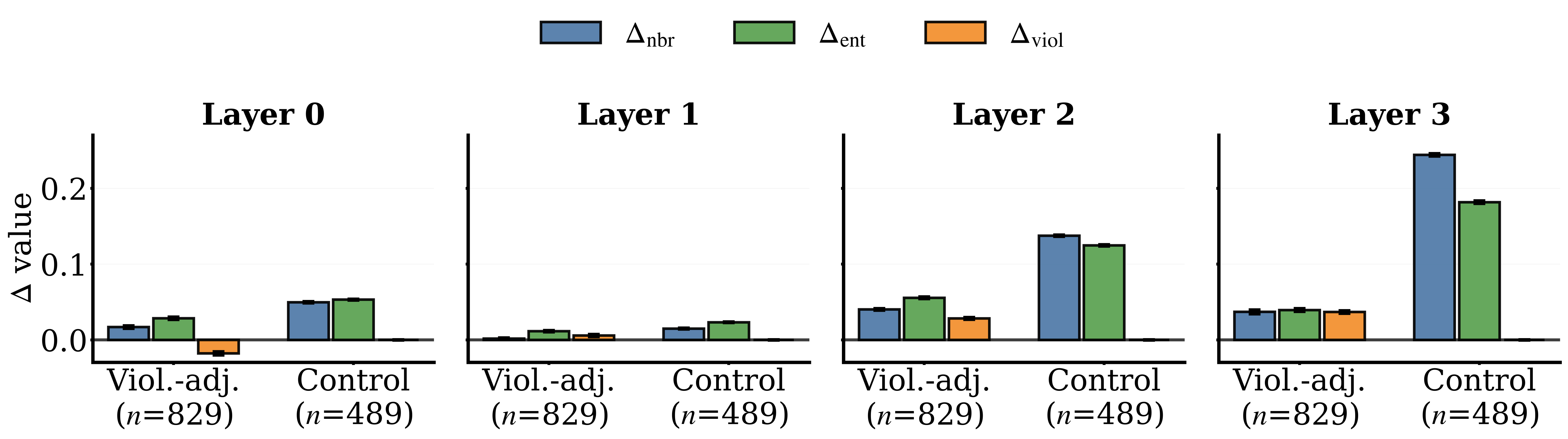}
    \caption{Sudoku L$-$H attention contrasts across query classes and Transformer layers.
    L-updates are more local than H-updates ($\Delta_{\mathrm{nbr}}, \Delta_{\mathrm{ent}} > 0$), while violation-specific concentration ($\Delta_{\mathrm{viol}} > 0$) appears mainly in deeper layers.
    $\Delta_{\mathrm{viol}}$ is zero on control queries by construction, because $\mathcal{N}(q)\cap\mathcal{V}=\emptyset$ for them.}
    \label{fig:attention_quant_summary}
\end{figure}

\begin{figure}[!htp]
    \centering
    \includegraphics[width=0.92\linewidth]{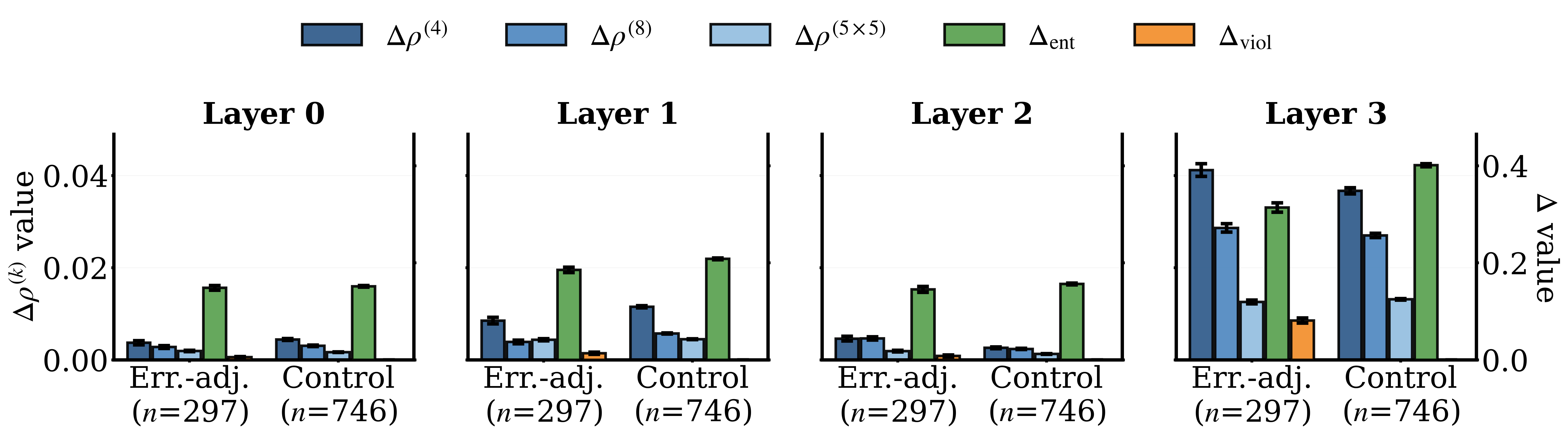}
    \caption{Maze attention contrasts across error-adjacent and control queries.
    We report per-cell density contrast $\Delta\rho^{(k)}$ for $k\in\{4, 8, 5{\times}5\}$, together with $\Delta_{\mathrm{ent}}$ and $\Delta_{\mathrm{viol}}$ over all four layers. $\Delta_{\mathrm{viol}}$ is zero on control queries by construction, because $\mathcal{N}_4(q)\cap\mathcal{V}=\emptyset$ for them.}
    \label{fig:maze_quant}
\end{figure}

\paragraph{L-updates are consistently more local than H-updates.}
On Sudoku, both $\Delta_{\mathrm{nbr}}$ and $\Delta_{\mathrm{ent}}$ are positive across query classes and layers (\Cref{fig:attention_quant_summary}).
Thus L places more attention inside $\mathcal{N}(q)$ and has a more concentrated distribution than H.
The gap is largest at the deepest layer: on control cells, $\Delta_{\mathrm{nbr}} = 0.24$, roughly $47\%$ more attention mass inside $\mathcal{N}(q)$ for L than for H.

Maze shows the same pattern under the size-invariant density measure.
For every neighborhood size $k$, layer, and query class, $\Delta\rho^{(k)}$ is positive (\Cref{fig:maze_quant}), meaning L places more attention near the query than H.
The effect is strongest for the smallest neighborhood and weakens as the neighborhood expands, indicating that L's extra attention is concentrated closest to the query.
The entropy contrast agrees: $\Delta_{\mathrm{ent}} > 0$ at every layer for both classes, so L attention is more concentrated over the full 900-cell grid.
Representative attention maps are in \Cref{app:attention_example,app:maze_example} for Sudoku and Maze respectively.

\paragraph{Violation-specific routing appears on Sudoku but not on Maze.}
On Sudoku, $\Delta_{\mathrm{viol}}$ is negative at layer~0 ($-0.018$), small at layer~1 ($+0.006$), and positive by layers~2--3 ($+0.028$ and $+0.037$).
Since L is already more local overall, this could reflect locality rather than sensitivity to violations.
To isolate violation sensitivity, we restrict attention to the same Sudoku neighborhood $\mathcal{N}(q)$ and ask whether L puts disproportionately more attention on violated cells than H does.
Specifically, within $\mathcal{N}(q)$, we compare the average attention per cell on violated cells, $\mathcal{N}(q)\cap\mathcal{V}$, against the average attention per cell on non-violated cells, $\mathcal{N}(q)\setminus\mathcal{V}$.
This control is near zero at layers~0--1 and positive at layers~2--3 ($+0.0025$ and $+0.0046$), showing that the deeper-layer signal exceeds locality alone.

On Maze, the analogous within-neighborhood control inside $\mathcal{N}_4(q)$ stays near zero at every layer.
Thus L's extra mass on Maze error cells is explained by locality, not by a separate violation-specific routing signal.

\paragraph{Causal attention interventions.}
To test whether attention locality causally contributes to solving, we additionally intervene on attention at inference across five independently trained checkpoints per task. These interventions show that L's neighborhood access supports solving on both tasks, while H's access beyond the neighborhood supports whole-maze solving. The protocols, controls, and results are provided in Appendix Sections \mbox{\ref{app:sudoku_causal_attention}} and \mbox{\ref{app:maze_causal_attention}} for Sudoku and Maze, respectively.
\Cref{app:interpretation} discusses possible explanations for these differences.

\section{Discussion and conclusion}\label{sec:conclusion}

This paper asked whether a shared-weight recurrent Transformer can develop distinct roles for two latent states when the only built-in asymmetry is which update receives the input.
On Sudoku-Extreme and Maze, the answer is yes: decoded rollouts show a committed $z_H$ and a revision-oriented $z_L$, L-updates attend more locally than H-updates, and freeze experiments show that the two states are dynamically interdependent.
Injection asymmetry also gives a consistent accuracy advantage on Sudoku, with more limited, configuration-dependent evidence on Maze.

Taken together, these results suggest that specialization does not require separate parameters.
What it requires is a clear signal of which state is being updated, which can come either from asymmetric input injection or from a structurally separable level token.

Looking ahead, stronger two-state recurrent reasoners with approximately constant-memory training, such as GRAM~\citep{baek2026gram} and Attractor Models~\citep{feinashley2026attractor}, are a natural place to test whether these findings carry over.
The same diagnostics---intermediate decoding, attention routing, and causal attention interventions---can check whether their states likewise differ in commitment, whether their input-receiving updates attend more locally, and whether these differences matter causally.

\Needspace{14\baselineskip}
\section{Limitations and broader impact}\label{sec:limitations}

All experiments are conducted on two synthetic, fully observable, grid-based combinatorial tasks, so our conclusions should be interpreted as evidence that asymmetric specialization can emerge in a controlled setting, not yet as a claim of universality.
We study only a 4-layer shared Transformer with two recurrent latent states and default $C_L = C_H = 2$ updates per cycle; our mechanistic evidence is strongest on Sudoku, and attention weights remain only a partial view of computation.
We train only with HRM's one-step gradient and have not tested whether state identity has the same effect under TRM-style backpropagation.
Whether the same role split persists under deeper blocks, different schedules, natural-language reasoning, continuous domains, or partial observability is left to future work.

This work is foundational research on the interpretability and inductive biases of small recurrent reasoning systems trained on synthetic tasks.
We do not foresee direct harmful societal impact from these models, though understanding recurrent specialization may inform more efficient and interpretable reasoning systems.

\begin{ack}
This research was funded in part by Rice University, the U.S. National Science Foundation (CAREER award no.~2145629), and a Microsoft Research Award. AK also thanks the Ken Kennedy Institute at Rice University for its support through the research cluster QuanTAS. We thank the anonymous reviewers and the area chair for comments that improved this paper. The authors declare no competing interests.
\end{ack}

\bibliographystyle{plainnat}

\bibliography{refs}

\clearpage
\appendix
\startcontents[appendix]
\phantomsection
\section*{Appendix}
\printcontents[appendix]{}{1}{\setcounter{tocdepth}{2}}
\clearpage
\section{Experimental details}\label{app:experimental_details}

\textbf{Architecture.}
All experiments use a single shared Transformer block with 4 layers, 8 attention heads, hidden dimension $D = 512$, and MLP expansion factor 4.
Rotary position embeddings (RoPE) are used.
The input encoder maps discrete tokens to $D$-dimensional embeddings.
The same output head (a linear projection followed by softmax over the token vocabulary) is used for both training supervision and intermediate-state decoding.
The output head is applied to $z_H$, so the training loss directly supervises $z_H$ only; $z_L$ is shaped indirectly through the recurrence.
Before the first cycle of each new input, $z_H$ and $z_L$ are initialized to broadcast copies of fixed vectors $\v{h}_0,\v{l}_0\in\R^D$ drawn once at construction from a truncated normal distribution with standard deviation $1$ (values outside $[-2, 2]$ resampled).

\textbf{Recurrence structure.}
Each high-level cycle interleaves $C_L$ L-updates with one H-update, repeated $C_H$ times, giving the pattern (L$\times C_L$, H)$\times C_H$.
With the default $C_L = C_H = 2$ this is \textbf{L, L, H, L, L, H}---six sub-steps per cycle.
The first four sub-steps run without gradient; gradient is computed only through the final L-update and the final H-update of each cycle (two updates), so that backpropagation traverses only the two last applications of the shared block in each cycle.
An ACT-style halting mechanism~\citep{graves2016adaptive} determines the number of high-level cycles per input, with a maximum of 16 cycles and a halt exploration probability of 0.1.

\textbf{Tasks.}
\emph{Sudoku-Extreme} uses $9 \times 9$ Sudoku puzzles with 17 given cells (the minimum for a unique solution), represented as a sequence of 81 tokens from a vocabulary of $\{1, \ldots, 9, \texttt{BLANK}\}$.
The \texttt{BLANK} token represents an unfilled cell.
\emph{Maze} uses $30 \times 30$ grid mazes, where the network predicts a token for each cell indicating wall, path, or solution.
The \texttt{PAD} token serves as the undecided marker analogous to \texttt{BLANK} in Sudoku.

\textbf{Training.}
Following the HRM training recipe~\citep{wang2025hrm}, all models are trained with AdamATan2~\citep{everett2024scaling} ($\beta_1 = 0.9$, $\beta_2 = 0.95$) using a learning rate of $1 \times 10^{-4}$, weight decay $1.0$, and global batch size $768$.
The learning rate warms up over 2{,}000 steps.
The loss function is stablemax cross-entropy~\citep{prieto2025grokking}, combined with HRM's Q-learning halting loss; both losses are computed from $z_H$.
The gradient update follows HRM's one-step truncated backpropagation~\citep{williams1990efficient}: the network performs every sub-step of the cycle, but the gradient is computed only through the final L-update and the final H-update (no unrolling through earlier sub-steps).
Sudoku models are trained for 20{,}000 epochs; Maze models use the same procedure.
Training data consists of 1{,}000 puzzles with 1{,}000$\times$ augmentation.
All accuracy numbers report peak / early-stop exact-match accuracy on a held-out test set.

\textbf{Error bars.}
Unless otherwise stated, evaluations use the full held-out test splits: 422{,}786 puzzles for Sudoku and 1{,}000 puzzles for Maze.
Unless otherwise stated, accuracy and ablation tables report mean $\pm$ standard deviation across 5 independent training runs with random seeds 0 through 4.
Freeze-experiment plots report mean $\pm$ sample standard deviation across three seeds, using 1{,}000 puzzles per seed (seeds 0--2, except the symmetric Maze control, which uses seeds 2--4).
Attention-analysis figures and tables report puzzle-level mean $\pm$ 95\% confidence intervals over the analyzed test puzzles.

\textbf{Compute.}
Each Sudoku training run takes approximately 4 hours on a single NVIDIA H200 GPU.
Each Maze training run takes approximately 3 hours on 8 NVIDIA H200 GPUs.
The total compute for the experiments reported in this paper, including ablations and preliminary runs, is approximately 500 GPU-hours.

\section{Significance tests for Table~\ref{tab:injection_maze}}\label{app:table1_significance}

\paragraph{Testing protocol.}
We compare each of the six asymmetric configurations in Table~\ref{tab:injection_maze} with each of the two symmetric configurations using two-sided Welch $t$-tests computed from the reported means and standard deviations, with $n=5$ independent training seeds per variant.
Each task defines a separate family of 12 comparisons, to which we apply Holm correction at $\alpha=0.05$.
Differences are asymmetric minus symmetric accuracy, in percentage points (pp).
The 95\% confidence intervals are individual Welch intervals, without multiplicity adjustment; significance is determined by the Holm-adjusted $p$-value.

\begin{table}[!htbp]
\centering
\caption{Asymmetric versus symmetric Sudoku variants from Table~\ref{tab:injection_maze}.}
\label{tab:sudoku_injection_significance}
\small
\setlength{\tabcolsep}{5pt}
\begin{tabular}{@{}llrcrrc@{}}
\toprule
Asymmetric & Symmetric & Difference & 95\% CI & $p$ & Holm $p$ & Significant? \\
\midrule
L\_Hx & Lx\_Hx & 6.6 & $[2.55,10.65]$ & 0.00748 & 0.0127 & Yes \\
L\_Hx & L2x\_H2x & 7.8 & $[3.26,12.34]$ & 0.00426 & 0.0127 & Yes \\
Lx\_H & Lx\_Hx & 7.9 & $[5.24,10.56]$ & 0.000156 & 0.00171 & Yes \\
Lx\_H & L2x\_H2x & 9.1 & $[5.39,12.81]$ & 0.000645 & 0.00580 & Yes \\
L\_H2x & Lx\_Hx & 6.5 & $[3.93,9.08]$ & 0.000426 & 0.00426 & Yes \\
L\_H2x & L2x\_H2x & 7.7 & $[4.02,11.38]$ & 0.00170 & 0.00986 & Yes \\
L2x\_H & Lx\_Hx & 7.0 & $[3.95,10.05]$ & 0.000995 & 0.00796 & Yes \\
L2x\_H & L2x\_H2x & 8.2 & $[4.29,12.11]$ & 0.00137 & 0.00957 & Yes \\
Lx\_H2x & Lx\_Hx & 7.5 & $[5.51,9.49]$ & 0.0000728 & 0.000874 & Yes \\
Lx\_H2x & L2x\_H2x & 8.7 & $[5.16,12.24]$ & 0.00164 & 0.00986 & Yes \\
L2x\_Hx & Lx\_Hx & 6.5 & $[2.91,10.09]$ & 0.00423 & 0.0127 & Yes \\
L2x\_Hx & L2x\_H2x & 7.7 & $[3.47,11.93]$ & 0.00300 & 0.0120 & Yes \\
\bottomrule
\end{tabular}
\end{table}
\FloatBarrier

\paragraph{Sudoku shows a consistent asymmetric-versus-symmetric separation.}
The 12 Sudoku comparisons in \mbox{\Cref{tab:sudoku_injection_significance}} yield positive differences of 6.5--9.1 percentage points, with all individual 95\% confidence intervals above zero.
All 12 comparisons remain significant after Holm correction within the Sudoku family (largest adjusted $p=0.0127$).
Thus, Sudoku provides statistically significant support for the asymmetric-versus-symmetric accuracy separation.

\begin{table}[!htbp]
\centering
\caption{Asymmetric versus symmetric Maze variants from Table~\ref{tab:injection_maze}.}
\label{tab:maze_injection_significance}
\small
\setlength{\tabcolsep}{4pt}
\begin{tabular}{@{}llrcrrc@{}}
\toprule
Asymmetric & Symmetric & Difference & 95\% CI & $p$ & Holm $p$ & Significant? \\
\midrule
L\_Hx & Lx\_Hx & 5.9 & $[1.67,10.13]$ & 0.0127 & 0.1267 & No \\
L\_Hx & L2x\_H2x & 5.0 & $[-0.51,10.51]$ & 0.0695 & 0.5561 & No \\
Lx\_H & Lx\_Hx & 1.6 & $[-6.09,9.29]$ & 0.6192 & 1.000 & No \\
Lx\_H & L2x\_H2x & 0.7 & $[-7.31,8.71]$ & 0.8421 & 1.000 & No \\
L\_H2x & Lx\_Hx & 6.2 & $[2.92,9.48]$ & 0.00265 & 0.0318 & Yes \\
L\_H2x & L2x\_H2x & 5.3 & $[0.16,10.44]$ & 0.0452 & 0.4069 & No \\
L2x\_H & Lx\_Hx & 1.7 & $[-6.11,9.51]$ & 0.6031 & 1.000 & No \\
L2x\_H & L2x\_H2x & 0.8 & $[-7.32,8.92]$ & 0.8220 & 1.000 & No \\
Lx\_H2x & Lx\_Hx & 1.5 & $[-2.07,5.07]$ & 0.3615 & 1.000 & No \\
Lx\_H2x & L2x\_H2x & 0.6 & $[-4.62,5.82]$ & 0.7903 & 1.000 & No \\
L2x\_Hx & Lx\_Hx & 5.1 & $[1.94,8.26]$ & 0.00671 & 0.0738 & No \\
L2x\_Hx & L2x\_H2x & 4.2 & $[-0.93,9.33]$ & 0.0895 & 0.6262 & No \\
\bottomrule
\end{tabular}
\end{table}
\FloatBarrier

\paragraph{Maze shows a positive average trend with configuration-dependent uncertainty.}
The descriptive group means favor asymmetric injection ($73.1\%$ versus $69.9\%$, a difference of approximately 3.2 percentage points), but the pairwise evidence is more limited.
Four of the 12 comparisons have unadjusted $p<0.05$; after Holm correction, only L\_H2x versus Lx\_Hx remains significant (difference 6.2 points, 95\% CI $[2.92,9.48]$, adjusted $p=0.0318$).
The largest uncertainty occurs for Lx\_H and L2x\_H, whose reported SDs are 6.3 and 6.4 points.
Thus, the higher average accuracy does not establish a uniform Maze advantage: the evidence depends on the configuration and its seed variability.

\FloatBarrier

\section{Level-token analysis}\label{app:level_token_analysis}
\subsection{Level-token interpretation}\label{app:level_token_interpretation}

The level-token control in \Cref{sec:injection} helps clarify what kind of state-identity signal AIR needs.
The failure of element-wise addition is informative.
Let $\tau_L,\tau_H \in \mathbb{R}^D$ denote the learned level vectors used to mark L- and H-updates, respectively, and let $\tau_m$ denote the vector for update type $m \in \{L,H\}$.
Suppose an L- or H-update receives token representations $X \in \mathbb{R}^{S \times D}$, where $S$ is the sequence length and $D$ is the hidden dimension.
Element-wise addition injects the level vector by replacing every token with
\[
    X_i' = X_i + \tau_m,\qquad m \in \{L,H\}.
\]
In a self-attention layer,
\[
    Q_i = X_i' W_Q,\qquad K_j = X_j' W_K,\qquad V_j = X_j' W_V .
\]
Because the same $\tau_m$ is added to every sequence position, it contributes the same state-dependent offsets
\[
    \tau_m W_Q,\qquad \tau_m W_K,\qquad \tau_m W_V
\]
to all queries, keys, and values.

Thus the level signal is not localized to a separate token or relation; it is mixed uniformly into every content token before attention is computed.

Expanding the attention logit between positions $i$ and $j$ makes the failure mode of the \emph{Addition} variant visible:
\[
    Q_i K_j^\top = X_i W_Q W_K^\top X_j^\top
        + X_i W_Q W_K^\top \tau_m^\top
        + \tau_m W_Q W_K^\top X_j^\top
        + \tau_m W_Q W_K^\top \tau_m^\top.
\]

Of the four terms above, only $\tau_m W_Q W_K^\top \tau_m^\top$ is a position-independent constant; the two cross-terms $X_i W_Q W_K^\top \tau_m^\top$ and $\tau_m W_Q W_K^\top X_j^\top$ depend on the position-varying token representations $X_i$ and $X_j$, so the contribution of $\tau_m$ to attention is content- and position-dependent rather than a uniform global bias.
What the shared Transformer still cannot do is read $\tau_m$ on its own.
At each position $i$, the query splits as $Q_i = X_i W_Q + \tau_m W_Q$.
The first part depends on the token $X_i$ and so varies by position.
The second part, $\tau_m W_Q$, is the same vector at every position.
The same split holds for the keys ($K_j = X_j W_K + \tau_m W_K$) and the values ($V_j = X_j W_V + \tau_m W_V$).
Because the level-vector terms $\tau_m W_Q$, $\tau_m W_K$, and $\tau_m W_V$ never appear on their own at any single position, an attention head has no way to read them in a structurally clean way. Prepend-and-strip avoids this problem.
There, the level token is at its own dedicated position, and attention can read $\tau_m W_*$ from that position without content mixed in.
This differs from asymmetric input injection, where the injected $\xemb$ is input-dependent and content-structured, and from prepend-and-strip, where the level signal occupies its own sequence position and can be selectively read through attention.

By contrast, prepend (strip) gives the model a structurally separable level token: the signal stays in its own sequence position, can be read through attention, and is removed after the call. Its value is also reset at every call, so it continues to mark the update type instead of accumulating content-dependent history. Prepend (no strip) instead allows its position-0 slot to accumulate recurrent content. The consistency analysis in \mbox{\Cref{app:level_token_training}} supports this interpretation, although its accuracy is not significantly different from prepend (strip) in the matched comparison.

Taken together, these controls support a transient, structurally separable state-identity signal as an effective design under the studied training setup; they do not establish that persistent tokens cannot provide useful identity information.

\subsection{Longer training and token consistency}\label{app:level_token_training}

\begin{figure}[!htbp]
\centering
\includegraphics[width=\linewidth]{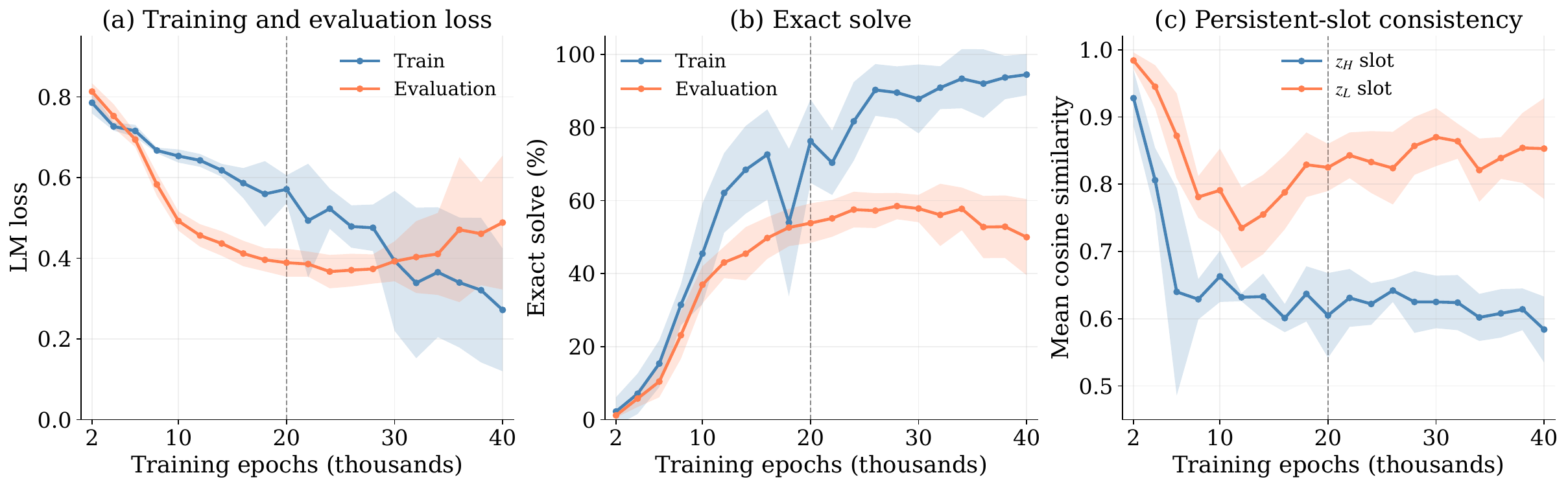}
\caption{Longer training of prepend (no strip) on Sudoku-Extreme with symmetric Lx\_Hx injection. The dashed line marks the main comparison's 20,000 budget.}
\label{fig:level_token_training}
\end{figure}

\paragraph{Training beyond the 20,000 budget.}
We train the symmetric Lx\_Hx prepend (no strip) control for five seeds to 40,000 epochs, twice the main comparison's budget, and record training and evaluation (using train and test split of the dataset, respectively) loss and exact solve every 2,000 epochs.
\mbox{\Cref{fig:level_token_training}} shows all 20 checkpoints as mean $\pm$ SD across seeds.
Unlike \mbox{\Cref{tab:level_tokens}}, which selects each seed's best evaluation within 20,000 epochs, these curves aggregate values at the same epoch across seeds.
This distinction accounts for the no-strip table value of $55.34\pm5.52\%$ versus $53.8\pm5.4\%$ at the fixed 20,000 checkpoint.

Training and evaluation diverge after the mid-training peak.
Training loss falls overall from $0.785\pm0.026$ at 2,000 to $0.272\pm0.152$ at 40,000, while training exact solve rises from $2.2\pm3.9\%$ to $94.5\pm5.7\%$.
Evaluation loss reaches its minimum at 24,000 ($0.367\pm0.041$), and evaluation exact solve peaks at 28,000 ($58.5\pm3.6\%$).
By 40,000, evaluation loss rises to $0.488\pm0.166$ and exact solve falls to $50.0\pm10.4\%$, below the fixed 20,000 value.
Thus, longer training gives a temporary gain followed by overfitting and greater variation across seeds.

\paragraph{Persistent slots acquire puzzle-dependent content.}
To examine whether the retained token remains a clean level marker, we measure its cross-puzzle consistency on 128 held-out puzzles at every checkpoint.
For each state $m\in\{H,L\}$, let $s_{m,i}$ denote the persistent position-0 vector for puzzle $i$ at the final recurrent step, and let $\bar{s}_m=128^{-1}\sum_{i=1}^{128}s_{m,i}$.
Consistency is the average cosine similarity $128^{-1}\sum_{i=1}^{128}\cos(s_{m,i},\bar{s}_m)$, computed separately for each seed.
A value near one indicates a common direction across puzzles; lower values indicate puzzle-dependent variation in the slot.

This variation develops early and persists throughout training.
H-slot consistency drops from $0.928\pm0.042$ at 2,000 to $0.640\pm0.154$ at 6,000, then fluctuates near $0.60$--$0.66$ before ending at $0.584\pm0.049$.
L-slot consistency falls from $0.984\pm0.012$ to $0.872\pm0.063$ over the same early interval and ends at $0.853\pm0.075$.
The learned level-vector norms at 40,000 are $\|\tau_H\|=0.584\pm0.037$ and $\|\tau_L\|=0.599\pm0.034$.
Thus, the level vectors learn nonzero values, while the recurrent slots that carry them acquire puzzle-specific content.
Longer training does not restore their initial cross-puzzle consistency.
These observations support content accumulation in persistent slots, but do not establish that it alone causes the later evaluation decline or that update identity becomes unrecoverable.
\FloatBarrier

\subsection{Pairwise tests for the level-token controls}\label{app:level_token_tests}

\Cref{tab:level_token_tests} reports all six pairwise comparisons among the four matched \texttt{Lx\_Hx} level-token controls in \Cref{tab:level_tokens}.
As in \Cref{app:table1_significance}, we use two-sided Welch $t$-tests over each run's best evaluation within 20{,}000 epochs ($n=5$ seeds per control) and apply Holm correction across the six comparisons at $\alpha=0.05$.
Differences are the first control minus the second, in percentage points.

\begin{table}[!htbp]
\centering
\caption{Pairwise tests among the matched \texttt{Lx\_Hx} level-token controls on Sudoku-Extreme (\Cref{tab:level_tokens}).}
\label{tab:level_token_tests}
\small
\setlength{\tabcolsep}{5pt}
\begin{tabular}{@{}lrcrrc@{}}
\toprule
Comparison & Difference & 95\% CI & $p$ & Holm $p$ & Significant? \\
\midrule
Addition $-$ Base       & $-1.16$ & $[-4.89, 2.57]$  & 0.478    & 0.788   & No  \\
Strip $-$ Base          & $6.16$  & $[3.85, 8.48]$   & 0.000281 & 0.00168 & Yes \\
Strip $-$ Addition      & $7.32$  & $[3.60, 11.05]$  & 0.00297  & 0.0149  & Yes \\
No-strip $-$ Base       & $3.27$  & $[-3.47, 10.02]$ & 0.263    & 0.788   & No  \\
No-strip $-$ Addition   & $4.43$  & $[-2.40, 11.26]$ & 0.165    & 0.659   & No  \\
Strip $-$ No-strip      & $2.89$  & $[-3.86, 9.64]$  & 0.314    & 0.788   & No  \\
\bottomrule
\end{tabular}
\end{table}
\FloatBarrier

\section{Operator-form control}\label{app:operator_forms}

\Cref{tab:operator_forms_app} reports accuracy on Sudoku-Extreme for four operator-form variants, all keeping the asymmetry pattern of the default Lx\_H configuration while replacing the default additive input $+\xemb$. The transformation $g$ in the linear and nonlinear variants is a learned $D \times D$ weight matrix that maps a token's $D$-dimensional embedding to another $D$-dimensional embedding, where $D$ is the shared Transformer's hidden dimension.
All four alternatives remain in the approximately $58$--$60\%$ interval, compared with $60.0\%$ for additive injection.
Two-sided Welch tests computed from the reported means and SDs ($n=5$ seeds per variant), with Holm correction across the four comparisons to the additive baseline, find no statistically detectable differences.
The differences range from $-0.1$ to $-1.7$ percentage points, and all individual 95\% confidence intervals include zero.
These results do not establish equivalence among operator forms.

\begin{table}[ht]
\centering
\caption{Operator-form control on Sudoku-Extreme.
Each variant modifies only the L-update argument while keeping the H-update as $f(z_H {+} z_L)$.
Mean $\pm$ standard deviation across 5 seeds.}
\label{tab:operator_forms_app}
\begin{tabular}{lcc}
\toprule
\textbf{Variant} & \textbf{L-input} & \textbf{Acc.\ (\%)} \\
\midrule
Default (additive) & $f(z_L {+} z_H {+} \xemb)$                       & $60.0_{\pm 2.0}$ \\
Linear transform    & $f(z_L {+} z_H {+} g(\xemb))$                    & $59.6_{\pm 1.3}$ \\
Nonlinear transform & $f(z_L {+} z_H {+} \tanh(g(\xemb)))$             & $59.9_{\pm 1.5}$ \\
Sign flip           & $f(z_L {+} z_H {-} \xemb)$                       & $59.2_{\pm 1.2}$ \\
Hadamard product    & $f(z_L {+} (z_H {+} z_L)\odot \xemb)$            & $58.3_{\pm 1.0}$ \\
\bottomrule
\end{tabular}
\end{table}
\section{Single-state ablation}\label{app:single_state}

The injection ablations in \Cref{sec:injection} show that the shared model needs a way to distinguish its two update types.
A natural follow-up question is whether the two latent states $z_H$ and $z_L$ are themselves necessary, or whether periodic input injection into a \emph{single} latent state would suffice.

To test this, we replace $z_H$ and $z_L$ with a single state $\v{z}$ and keep everything else identical.
Within the original L-cycle, $\v{z}$ receives input injection ($\v{z} \leftarrow f(\v{z} + \xemb)$); within the original H-cycle, it does not ($\v{z} \leftarrow f(\v{z})$). This is analogous to the L,L,H,L,L,H pattern used in the default two-states asymmetric variant.

\begin{table}[ht]
\centering
\caption{Single-state ablation on Sudoku-Extreme (5 seeds).}
\label{tab:single_state}
\begin{tabular}{lc}
\toprule
\textbf{Variant} & \textbf{Acc.\ (\%)} \\
\midrule
Single state $\v{z}$ (periodic injection) & $51.1_{\pm 1.8}$ \\
\midrule
\emph{For reference:} \\
\quad Symmetric two-state, $(1,1)$ & $52.1_{\pm 1.6}$ \\
\quad Asymmetric two-state, $(1,0)$ (default) & $60.0_{\pm 2.0}$ \\
\bottomrule
\end{tabular}
\end{table}

The single-state model achieves $51.1\%$ accuracy, on par with the symmetric two-states baselines and far below the asymmetric default ($60.0\%$).
This indicates that periodic input injection alone is not sufficient.
A useful way to see the difference is to write the single-state recurrence as one stream
\[
    \v{z}^{(t+1)} =
    f\!\left(\v{z}^{(t)} + a_t \xemb;\param\right),
    \qquad a_t \in \{0,1\},
\]
where $a_t$ is a periodic injection indicator.
Although the shared model can observe whether the current sub-step is injected or non-injected, both update types overwrite the same state $\v{z}$.
Thus any injected-step features and non-injected-step features must coexist in the same representational buffer:
\[
    \v{z}^{(t)}
    =
    \Phi_t(\v{z}^{(0)}, \xemb; a_0,\ldots,a_{t-1}).
\]
By contrast, AIR maintains two persistent slots,
\[
    z_L^{(t+1)} =
    f\!\left(z_L^{(t)} + z_H^{(t)} + \xemb;\param\right),
    \qquad
    z_H^{(t+1)} =
    f\!\left(z_H^{(t)} + z_L^{(t+1)};\param\right).
\]
The injected and non-injected computations are therefore not merely two phases of one trajectory; they are written back into different latent states.
This gives the model two memory buffers with different update histories: $z_L$ accumulates the input-conditioned, partially committed refinement dynamics, while $z_H$ accumulates the non-injected, fully committed proposal dynamics.
The single-state control removes this storage separation, so the injected and non-injected states cannot stably specialize into separate latent roles.

\clearpage
\section{Attention analysis on Sudoku}\label{app:attention_viz}

This section gives the formal definitions for the Sudoku attention statistics used in \Cref{sec:mechanism} and provides additional visualizations. We focus on the same L-update versus H-update comparison as in \Cref{sec:mechanism}.

\subsection{Setup}

\textbf{Notation.}
Throughout this appendix, we identify each test puzzle by a tag \texttt{pNNNN} (e.g., \texttt{p0121}) and denote a board cell by \(rXcY\), meaning the cell at row $X$ and column $Y$ of the $9{\times}9$ Sudoku board (1-indexed, with \(r1c1\) at the top-left and \(r9c9\) at the bottom-right).

\textbf{Formal definitions.}
For each blank query cell $q$ and attention head $h$, let $\tilde p_q^{(L)}(h), \tilde p_q^{(H)}(h) \in \mathbb{R}_{\ge 0}^{81}$ denote the raw per-head attention weights from $q$ to the 81 puzzle-cell keys at the L-update and H-update.
For each query cell we report the arithmetic mean across the 8 heads of
\begin{align*}
\Delta_{\mathrm{nbr}}(q) &= \mathrm{mass}_{L}^{\mathrm{nbr}}(q) - \mathrm{mass}_{H}^{\mathrm{nbr}}(q), \\
\Delta_{\mathrm{viol}}(q) &= \mathrm{mass}_{L}^{\mathrm{viol}}(q) - \mathrm{mass}_{H}^{\mathrm{viol}}(q), \\
\Delta_{\mathrm{ent}}(q) &= \mathrm{entropy}(p_q^{(H)}) - \mathrm{entropy}(p_q^{(L)}),
\end{align*}
where $\mathcal{N}(q)$ is the Sudoku neighborhood of a query cell $q$ (the 20 cells sharing a row, column, or $3{\times}3$ box) and $\mathcal{V}$ is the set of cells currently violating a constraint under decoded $z_H$.
Let $k$ denote the key cell to which $q$ attends. Then $\mathrm{mass}^{\mathrm{nbr}}(q) = \sum_{k \in \mathcal{N}(q)} \tilde p_{qk}$, $\mathrm{mass}^{\mathrm{viol}}(q) = \sum_{k \in \mathcal{N}(q) \cap \mathcal{V}} \tilde p_{qk}$, and $\mathrm{entropy}(p_q) = -\sum_k p_{qk} \log p_{qk} \,/\, \log 81$ over the normalized distribution $p_{qk} = \tilde p_{qk}/\sum_{k'}\tilde p_{qk'}$.

\textbf{Within-neighborhood control.}
For a subset $S \subseteq \mathcal{N}(q)$ and update type $m \in \{L, H\}$, define the per-cell attention density $\rho_m^S(q) = \tfrac{1}{|S|}\sum_{k \in S} \tilde p_{qk}^{(m)}$.
The within-neighborhood control statistic used in \Cref{sec:mechanism} is then
\[
    \Delta\rho_{\mathrm{viol}}(q)
    \;=\; \bigl[\rho_L^{\mathcal{N}(q)\cap\mathcal{V}}(q) - \rho_H^{\mathcal{N}(q)\cap\mathcal{V}}(q)\bigr]
    \;-\; \bigl[\rho_L^{\mathcal{N}(q)\setminus\mathcal{V}}(q) - \rho_H^{\mathcal{N}(q)\setminus\mathcal{V}}(q)\bigr],
\]
that is, the L$-$H per-cell density on violated cells inside $\mathcal{N}(q)$ minus the L$-$H per-cell density on the non-violated cells inside $\mathcal{N}(q)$. We restrict the average to violation-adjacent queries whose $\mathcal{N}(q)$ contains both bins (so both $\rho^{\mathcal{N}(q)\cap\mathcal{V}}$ and $\rho^{\mathcal{N}(q)\setminus\mathcal{V}}$ are well defined). The Maze appendix (\Cref{app:maze_attention}) uses the analogous statistic with $\mathcal{N}(q)$ replaced by $\mathcal{N}_4(q)$.

\subsection{Representative examples}\label{app:attention_example}

\begin{figure}[!htbp]
    \centering
    \begin{subfigure}{\linewidth}
        \centering
        \includegraphics[width=0.85\linewidth]{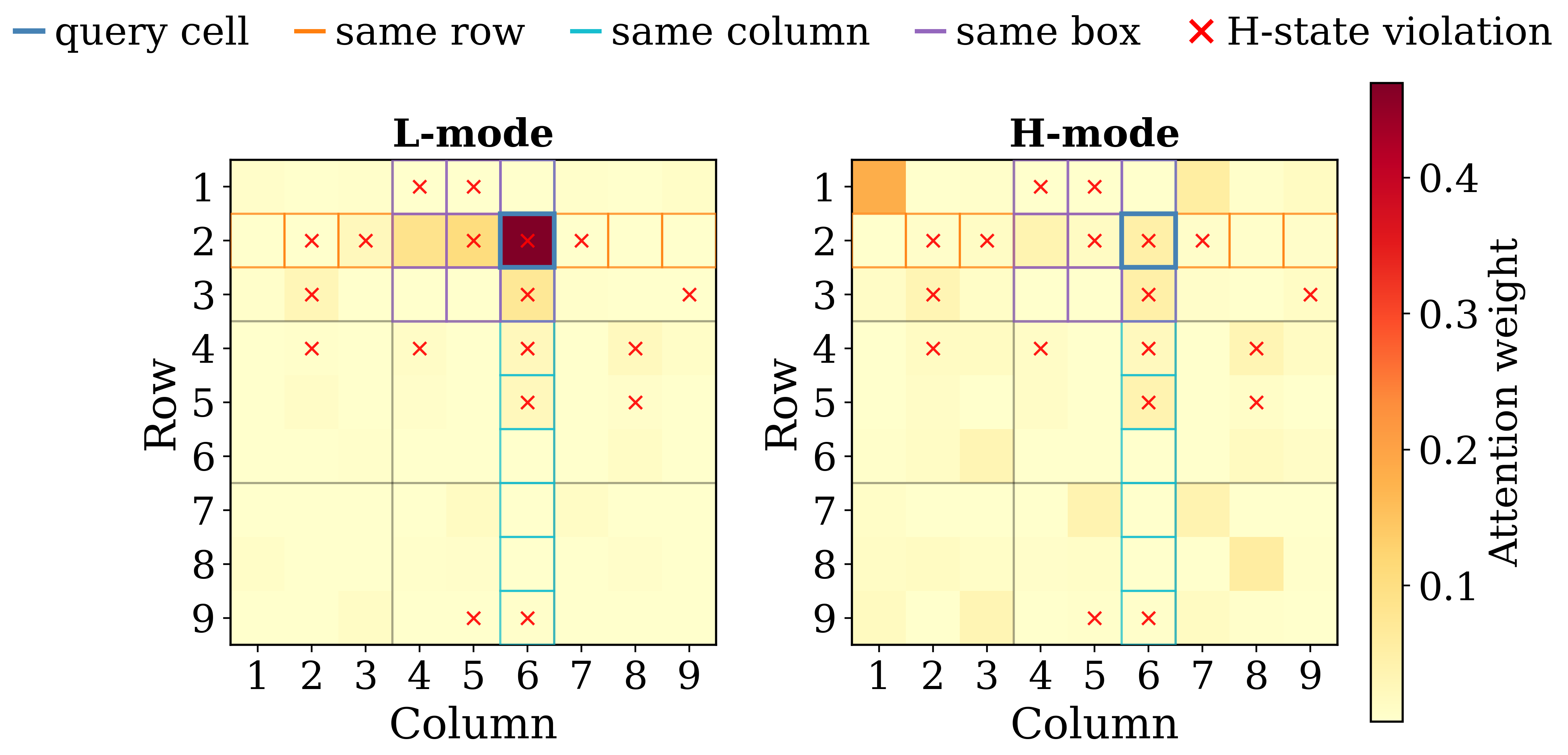}
        \caption{Puzzle \texttt{p0121}, cycle~2, head~1, layer~0, query \(r2c6\). L-update concentrates roughly $0.81$ of its mass inside $\mathcal{N}(q)$ versus $0.24$ for H-update.}
        \label{fig:attention_core_example_app}
    \end{subfigure}
    \\[0.6em]
    \begin{subfigure}{\linewidth}
        \centering
        \includegraphics[width=0.85\linewidth]{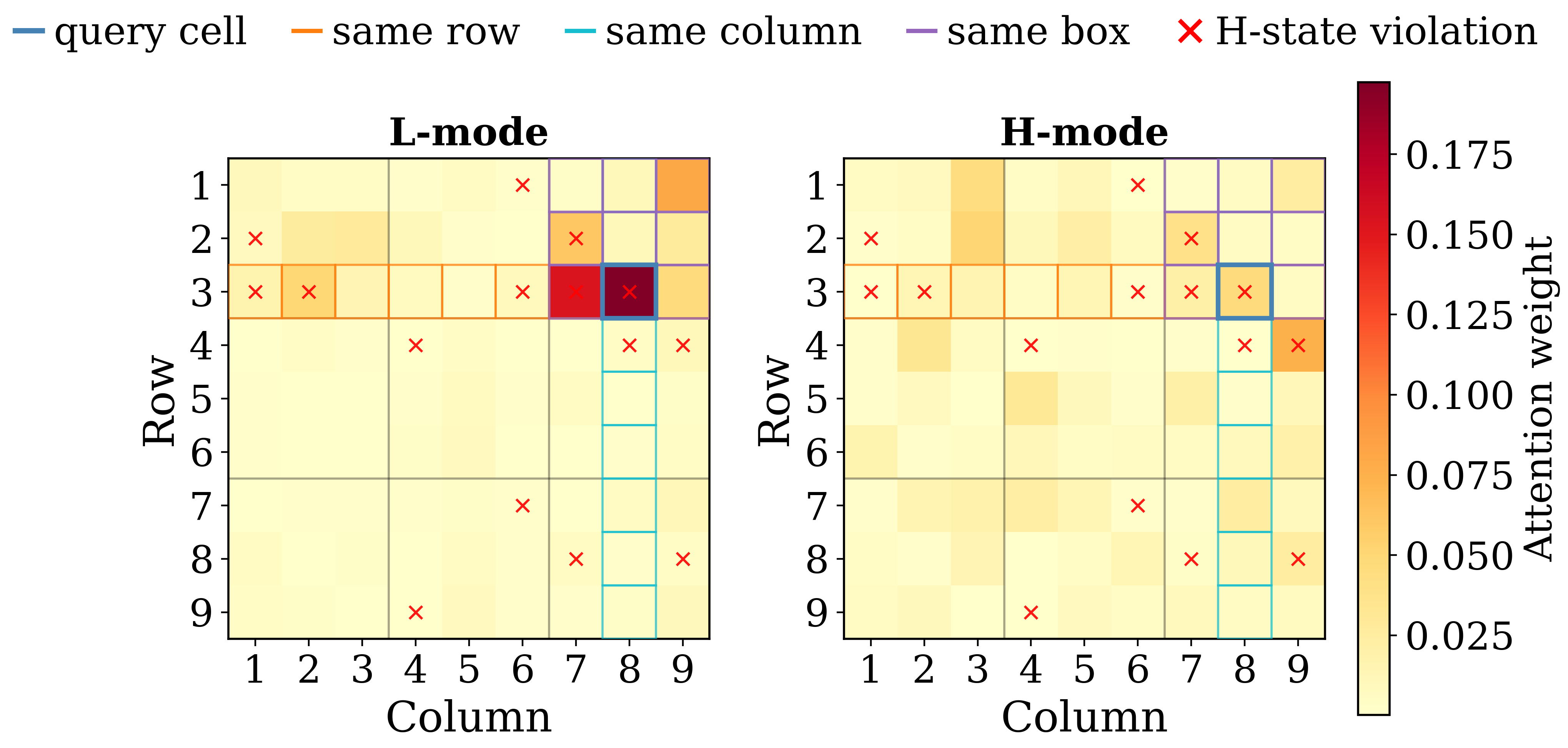}
        \caption{Puzzle \texttt{p0926}, cycle~2, head~0, layer~0, query \(r3c8\). L-update concentrates most of its attention onto the same-box neighbor (e.g.\ cells \(r3c7\), \(r1c9\)), while H-update places substantial mass outside the neighborhood (e.g., \(r4c9\)).}
        \label{fig:attention_core_example_app2}
    \end{subfigure}
    \\[0.6em]
    \begin{subfigure}{\linewidth}
        \centering
        \includegraphics[width=0.85\linewidth]{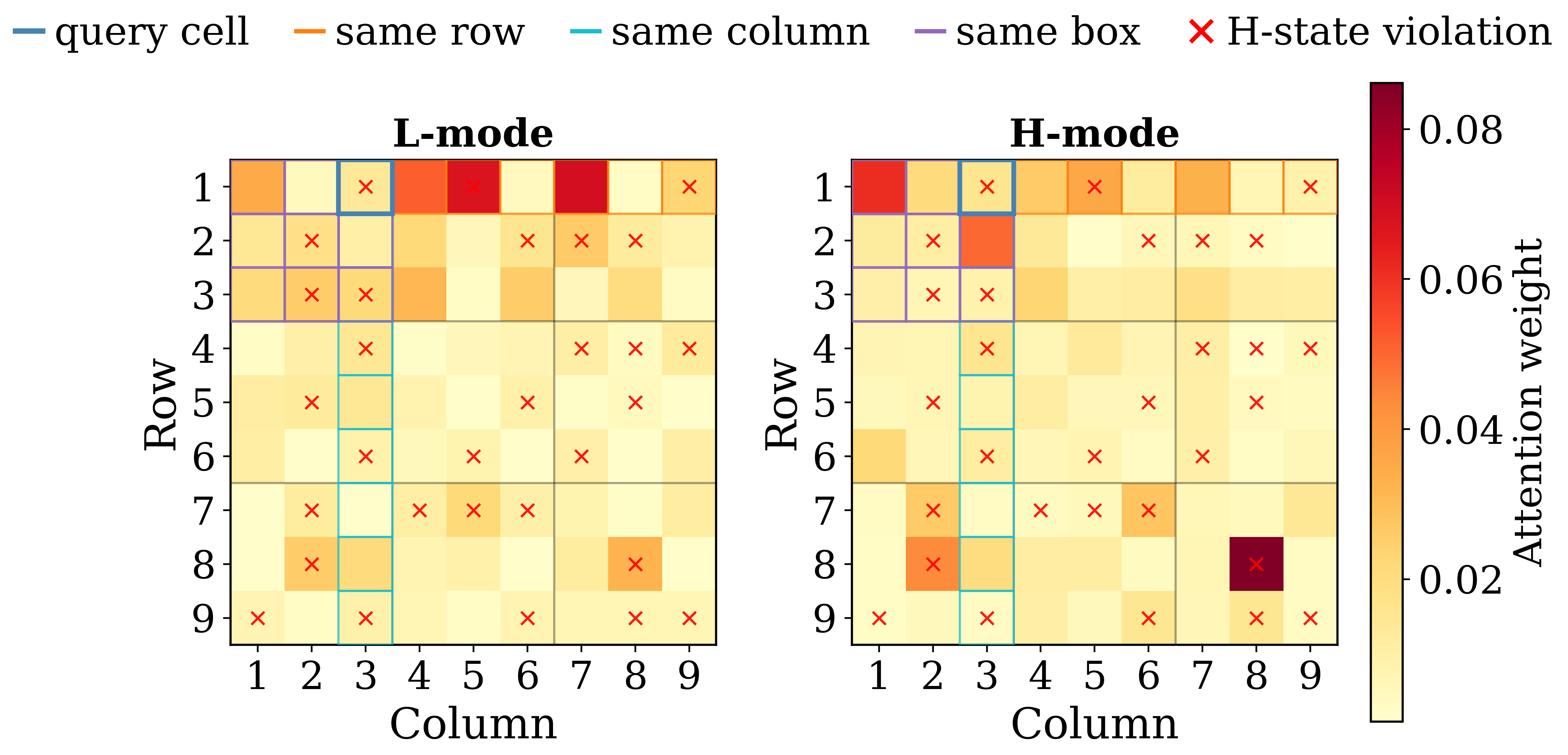}
        \caption{Puzzle \texttt{p0025}, cycle~4, head~2, layer~0, query \(r1c3\). L-update concentrates along the query's row inside the neighborhood; H-update is more global.}
        \label{fig:attention_core_example_app3}
    \end{subfigure}
    \caption{Three representative L-update versus H-update attention examples on Sudoku.}
    \label{fig:attention_core_examples}
\end{figure}

\Cref{fig:attention_core_example_app} shows a blank query cell whose neighborhood contains a violated cell (puzzle \texttt{p0121}, cycle~2, head~1, layer~0). For this query, L-update is more concentrated inside $\mathcal{N}(q)$, places more attention mass on violating cells, and has lower entropy than H-update. \Cref{fig:attention_core_example_app2,fig:attention_core_example_app3} show two additional queries from different puzzles and cycles with similar qualitative patterns.

\subsection{Temporal persistence}\label{app:attention_temporal}

\begin{figure}[!htbp]
    \centering
    \includegraphics[width=\linewidth]{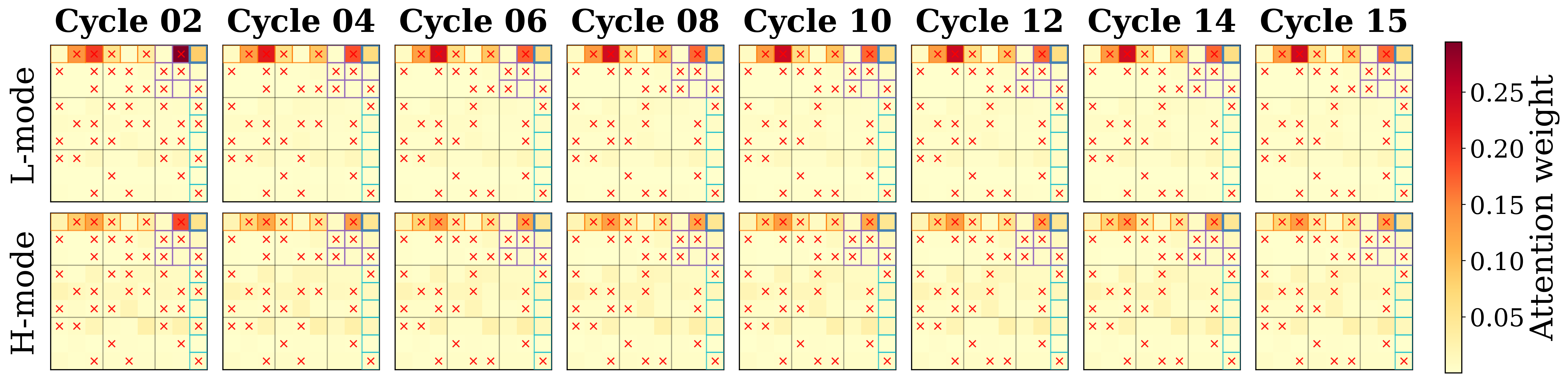}
    \caption{Temporal persistence for a single query cell (puzzle \texttt{p0829}, head~5, layer~0, query \(r1c9\)) across cycles~2 through~15.}
    \label{fig:attention_temporal_example_app}
\end{figure}

\Cref{fig:attention_temporal_example_app} tracks one query cell, head, and layer across multiple cycles. The L-update/H-update separation in terms of attention mass distribution persists over time, which we denote as \textit{temporal persistence}: L-update repeatedly emphasizes neighboring regions, whereas H-update remains more global. This rules out the interpretation that the difference is confined to a single cycle.

\FloatBarrier
\subsection{Layer robustness}\label{app:attention_layer_robustness}

\Cref{fig:attention_quant_summary} in \Cref{sec:mechanism} shows the L-update/H-update separation across all four layers; \Cref{tab:layer_robustness} gives the corresponding numerical values with 95\% confidence intervals.

\begin{table}[!htbp]
\centering
\caption{L-update/H-update attention separation across all Transformer layers, averaged over the 8 heads at each layer (puzzle-level mean $\pm$ 95\% CI). Aggregated over $n{=}829$ violation-adjacent and $n{=}489$ control puzzles drawn from 1{,}000 test puzzles. $\Delta_{\mathrm{viol}}$ is zero on control cells by construction, because $\mathcal{N}(q)\cap\mathcal{V}=\emptyset$ for them.}
\label{tab:layer_robustness}
\begin{tabular}{llccc}
\toprule
\textbf{Class} & \textbf{Layer} & $\Delta_{\mathrm{nbr}}$ & $\Delta_{\mathrm{ent}}$ & $\Delta_{\mathrm{viol}}$ \\
\midrule
Viol.-adj. & 0 & $0.017_{\pm 0.002}$ & $0.029_{\pm 0.001}$ & $-0.018_{\pm 0.002}$ \\
           & 1 & $0.002_{\pm 0.001}$ & $0.012_{\pm 0.001}$ & $0.006_{\pm 0.001}$ \\
           & 2 & $0.040_{\pm 0.001}$ & $0.056_{\pm 0.001}$ & $0.028_{\pm 0.001}$ \\
           & 3 & $0.037_{\pm 0.002}$ & $0.039_{\pm 0.002}$ & $0.037_{\pm 0.002}$ \\
\midrule
Control    & 0 & $0.050_{\pm 0.001}$ & $0.053_{\pm 0.000}$ & $0.000_{\pm 0.000}$ \\
           & 1 & $0.015_{\pm 0.001}$ & $0.023_{\pm 0.000}$ & $0.000_{\pm 0.000}$ \\
           & 2 & $0.138_{\pm 0.001}$ & $0.125_{\pm 0.001}$ & $0.000_{\pm 0.000}$ \\
           & 3 & $0.244_{\pm 0.002}$ & $0.182_{\pm 0.002}$ & $0.000_{\pm 0.000}$ \\
\bottomrule
\end{tabular}
\end{table}

The local-versus-global split is visible throughout the network: $\Delta_{\mathrm{nbr}} > 0$ and $\Delta_{\mathrm{ent}} > 0$ at every layer for control cells, and at all layers for violation-adjacent cells as well, though the effect is weakest at layer~1. The violation-specific signal behaves differently. At layer~0, $\Delta_{\mathrm{viol}}$ is slightly negative on violation-adjacent cells, so L-updates are not yet preferentially routing attention to error cells within the neighborhood. By layers~2 and~3, however, $\Delta_{\mathrm{viol}}$ is clearly positive. We therefore interpret locality as an early and robust property of L-updates, with explicit violation sensitivity emerging later in depth.

\FloatBarrier
\subsection{Causal attention interventions}\label{app:sudoku_causal_attention}

\paragraph{Intervention protocol.}
The locality analysis in Section 7 describes where L and H attend; here we test whether changing that access affects solving.
We intervene on attention at inference, keeping model parameters fixed, and evaluate five independently trained checkpoints on the same 1,000 Sudoku test puzzles.
Because the answer is decoded only from the H-state, intervening on the final H-update would directly alter the update that produces the decoded output.
We therefore leave the final H-update of each recurrent cycle unchanged and intervene only on its earlier H-updates; L interventions affect every L-update.

We use two interventions: attention knockout and diffuse attention.
Knockout removes selected keys from attention by adding a large negative mask ($-10^9$) to their logits before softmax.
For a query cell $q$, let $\mathcal{N}(q)$ be its 20 peers sharing a row, column, or $3\times3$ box.
We either forbid access to $\mathcal{N}(q)$ or confine access to it, calling the latter a local update.
To control for the number of accessible cells, we also confine attention to a random set of 20 other cells per query, sampled once with a fixed seed and reused across checkpoints.
All knockout masks retain self-attention and unrestricted access to and from puzzle-prefix tokens.
A diffuse update instead spreads attention uniformly over all sequence positions, including prefix tokens.
Both interventions apply to every head and layer of the selected update type.

\paragraph{Evaluation.}
We measure per-cell accuracy and exact solve (the percentage of puzzles with every evaluated cell correct) from the final decoded H-state after the normal halting procedure.
\mbox{\Cref{tab:sudoku_causal_attention}} reports mean $\pm$ sample SD across five seeds; changes below are relative to the no-intervention baseline in this table and are expressed in percentage points.
We test effects with Welch $t$-tests on seed-level measurements.
A one-sided test is used only for a disruption predicted a priori to degrade performance relative to baseline; comparisons between interventions and other comparisons use two-sided tests.

\begin{table}[!htbp]
\centering
\caption{Sudoku causal attention interventions at inference (five-seed mean $\pm$ SD).}
\label{tab:sudoku_causal_attention}
\begin{tabular}{@{}lrr@{}}
\toprule
Intervention & Per-cell accuracy & Exact solve \\
\midrule
None (baseline) & $83.6_{\pm 0.6}$ & $55.2_{\pm 1.6}$ \\
L diffuse & $51.2_{\pm 3.8}$ & $0.0_{\pm 0.0}$ \\
L excludes neighborhood & $53.7_{\pm 6.1}$ & $0.0_{\pm 0.1}$ \\
L local only & $64.2_{\pm 9.9}$ & $12.8_{\pm 13.8}$ \\
L random only & $51.1_{\pm 8.2}$ & $0.0_{\pm 0.0}$ \\
H diffuse & $78.5_{\pm 1.4}$ & $36.9_{\pm 5.0}$ \\
H excludes neighborhood & $59.8_{\pm 1.6}$ & $0.8_{\pm 0.7}$ \\
H local only & $81.7_{\pm 1.5}$ & $51.2_{\pm 4.0}$ \\
L local, H diffuse & $60.4_{\pm 5.6}$ & $4.4_{\pm 6.8}$ \\
L diffuse, H local & $57.2_{\pm 1.9}$ & $0.1_{\pm 0.2}$ \\
Both diffuse & $47.3_{\pm 2.3}$ & $0.0_{\pm 0.0}$ \\
\bottomrule
\end{tabular}
\end{table}

\paragraph{L's neighborhood access contributes causally to solving.}
Making L diffuse reduces per-cell accuracy by 32.4 percentage points ($p=2\times10^{-5}$), and excluding its neighborhood reduces it by 29.9 points ($p=2\times10^{-4}$); both nearly eliminate exact solving.
These interventions change attention while leaving the learned parameters fixed, showing that L's access to neighboring cells is part of the computation supporting correct answers.
Confining L to its neighborhood retains some exact solving ($12.8\%$), but remains below baseline ($55.2\%$), indicating that local access alone is insufficient to preserve performance.

\paragraph{Controlling which cells L can attend to.}
To distinguish neighborhood structure from the number of accessible cells, we compare local attention with the matched-size random control.
Both allow L to attend to 20 other cells and differ only in their locations.
The true neighborhood retains more per-cell accuracy ($64.2\%$ versus $51.1\%$), corresponding to reductions of 19.4 and 32.5 points from baseline.
This ordering suggests that the choice of cells matters, although the Sudoku comparison does not reach the $0.05$ significance threshold ($p=0.053$).

\paragraph{Disrupting L attention increases rule violations per error.}
Per-cell accuracy alone does not reveal whether a prediction respects Sudoku constraints: a cell can differ from the target solution while remaining consistent with its row, column, and box.
We therefore also measure violations per wrong cell in the final decoded H-state.
A filled cell participates in a violation if its value repeats another in its row, column, or box; each participating cell is counted once.
For each seed, we divide the total number of these cells by the total number of cells differing from the target solution.
This ratio can exceed one because correctly predicted cells can also conflict with incorrect neighbors.

Making L diffuse raises violations per wrong cell from $0.84$ to $1.41$ ($p<10^{-8}$; \mbox{\Cref{tab:sudoku_causal_violations}}), showing that the disruption increases rule conflicts relative to the number of prediction errors.
Making H diffuse gives a ratio near baseline ($0.91$, $p=0.72$), with no statistically detectable change.
Together, these results support a role for L's attention in maintaining local constraint consistency.

\begin{table}[!htbp]
\centering
\caption{Sudoku rule violations under attention interventions (five-seed mean $\pm$ SD).}
\label{tab:sudoku_causal_violations}
\begin{tabular}{@{}lrr@{}}
\toprule
Intervention & Per-cell accuracy (\%) & Violations per wrong cell \\
\midrule
None (baseline) & $83.6_{\pm 0.6}$ & $0.84_{\pm 0.04}$ \\
L diffuse & $51.2_{\pm 3.8}$ & $1.41_{\pm 0.03}$ \\
H diffuse & $78.5_{\pm 1.4}$ & $0.91_{\pm 0.38}$ \\
\bottomrule
\end{tabular}
\end{table}
\FloatBarrier

\section{Attention analysis on Maze}
\label{app:maze_attention}

This section gives the formal definitions for the Maze attention statistics used in \Cref{sec:mechanism} and provides the supporting visualizations.
As in \Cref{sec:mechanism}, \emph{local} means attention concentrated near the query, \emph{global} means more global attention, and all reported quantities are averaged over the 8 attention heads.

\subsection{Setup}

\textbf{Notation.}
We identify each Maze-30$\times$30 test puzzle by a tag \texttt{pNNNN} and denote a board cell by \(rXcY\), the cell at row $X$ and column $Y$ of the $30{\times}30$ grid (1-indexed, with \(r1c1\) at the top-left and \(r30c30\) at the bottom-right). All maze panels show an $11{\times}11$ window centered on the query.

We extract attention from all four layers and compare the first L-update with the first H-update across cycles $\{2,4,6,8,10,12,14,15\}$. We take the cells that lie in the solution path in the $30{\times}30$ maze as query cells. For each query cell $q$, we define three neighborhoods analogous to the Sudoku neighborhood: the 4-connected neighborhood $\mathcal{N}_4(q)$, the 8-connected neighborhood $\mathcal{N}_8(q)$, and the centered $5{\times}5$ window $\mathcal{N}_{5\times5}(q)$ (excluding $q$ itself), with $\mathcal{N}_4(q) \subseteq \mathcal{N}_8(q) \subseteq \mathcal{N}_{5\times5}(q)$.

The violation set $\mathcal{V}$ contains cells where decoded $z_H$ disagrees with the ground truth. A query is \emph{error-adjacent} if $\mathcal{N}_4(q) \cap \mathcal{V} \neq \emptyset$ and \emph{control} otherwise. For each query we report the per-cell density contrast
\[
\Delta\rho^{(k)}(q) = \frac{\mathrm{mass}_L(\mathcal{N}_k(q)) - \mathrm{mass}_H(\mathcal{N}_k(q))}{|\mathcal{N}_k(q)|},
\]
which keeps comparisons across neighborhood sizes on the same scale; the entropy contrast $\Delta_{\mathrm{ent}}(q) = \mathrm{entropy}(p_q^{(H)}) - \mathrm{entropy}(p_q^{(L)})$ where $\mathrm{entropy}(p_q) = -\sum_k p_{qk} \log p_{qk} \,/\, \log 900$ over the normalized 900-cell distribution; and the violation contrast $\Delta_{\mathrm{viol}}(q)$, the L--H difference in mass on $\mathcal{N}_4(q) \cap \mathcal{V}$. We aggregate over 1{,}000 test puzzles, yielding $n{=}297$ error-adjacent and $n{=}746$ control puzzles after class assignment of error-adjacent and control, and report 95\% confidence intervals.

\subsection{Representative examples}\label{app:maze_example}

\begin{figure}[!htbp]
    \centering
    \begin{subfigure}{\linewidth}
        \centering
        \includegraphics[width=0.85\linewidth]{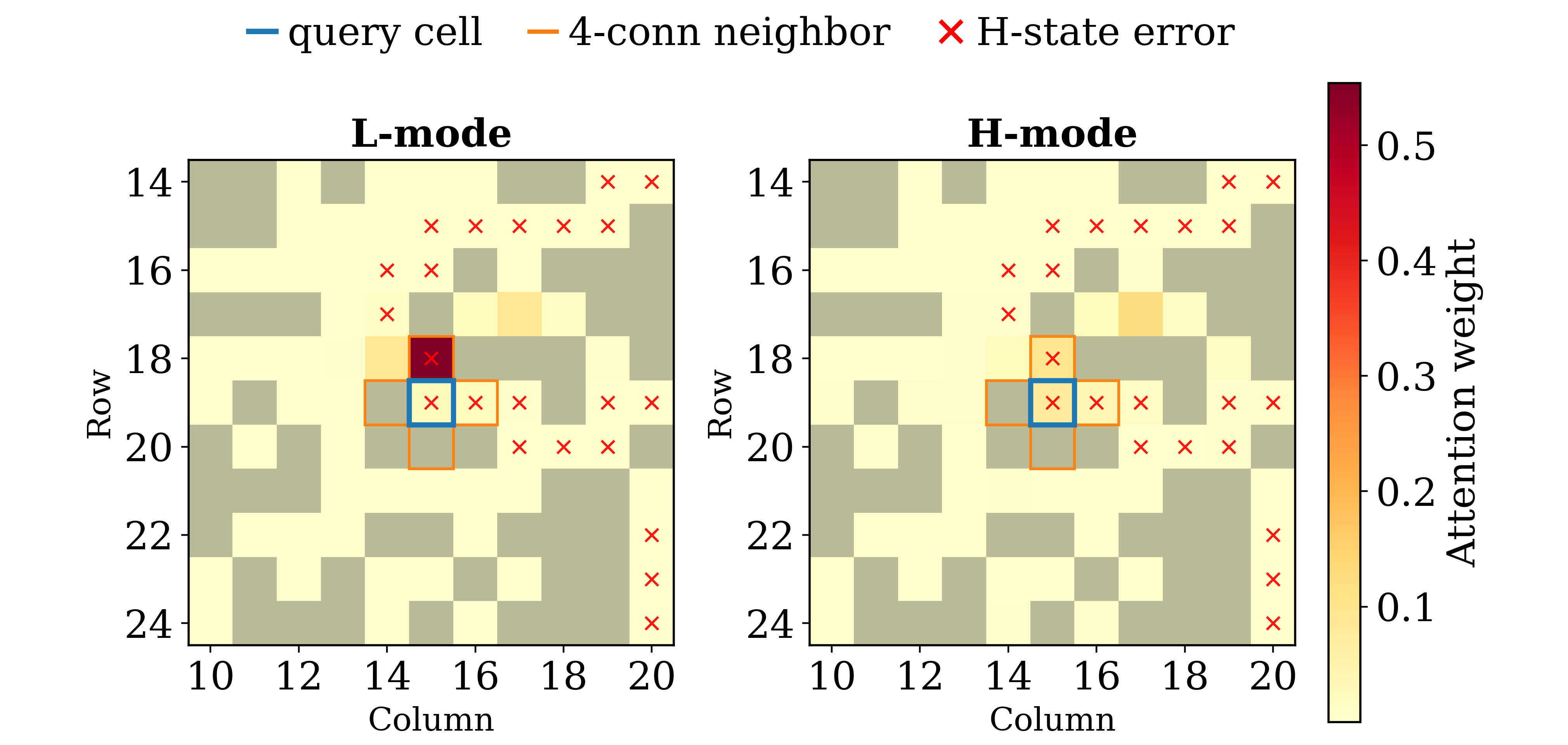}
        \caption{Puzzle \texttt{p0040}, cycle~4, head~7, layer~0, query \(r19c15\). L-update places almost all of its attention mass inside $\mathcal{N}_4(q)$; H-update spreads that mass over a broader region.}
        \label{fig:maze_core_example_app}
    \end{subfigure}
    \\[0.6em]
    \begin{subfigure}{\linewidth}
        \centering
        \includegraphics[width=0.85\linewidth]{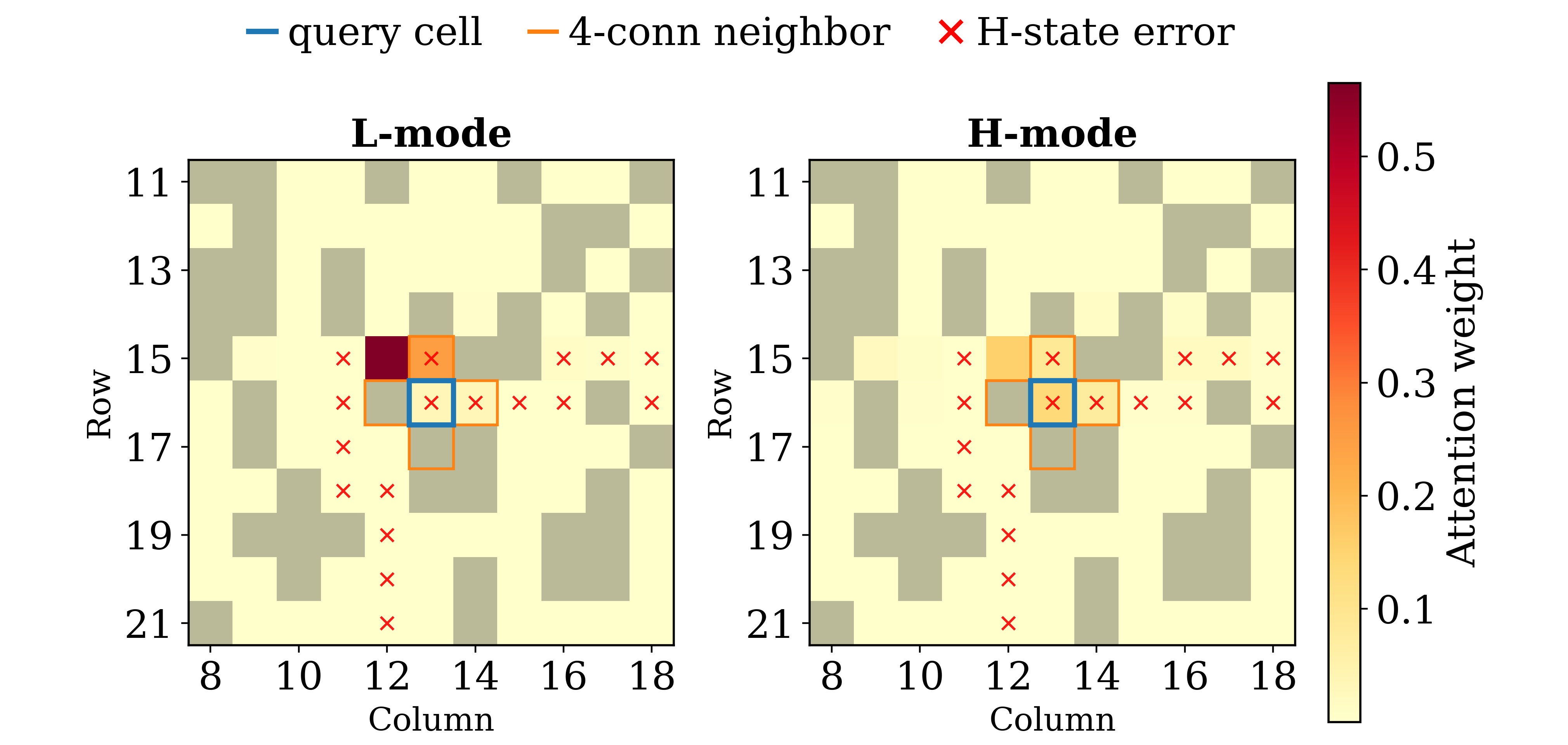}
        \caption{Puzzle \texttt{p0029}, cycle~12, head~7, layer~0, query \(r16c13\). L-update remains sharply local; H-update is broader and less peaked.}
        \label{fig:maze_core_example_app2}
    \end{subfigure}
    \\[0.6em]
    \begin{subfigure}{\linewidth}
        \centering
        \includegraphics[width=0.85\linewidth]{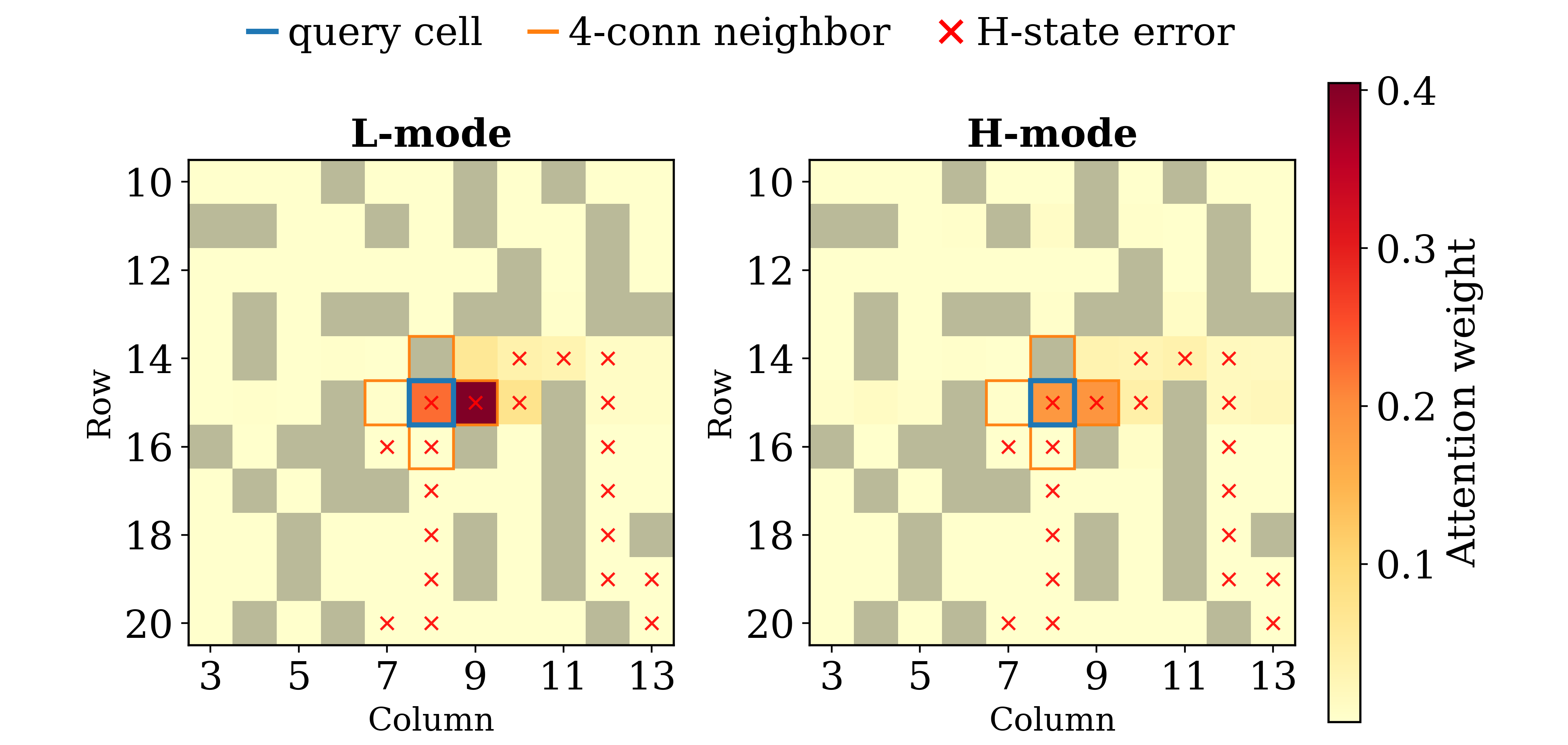}
        \caption{Puzzle \texttt{p0016}, cycle~12, head~7, layer~0, query \(r15c8\). The same local-versus-global split appears again.}
        \label{fig:maze_core_example_app3}
    \end{subfigure}
    \caption{Three representative L-update versus H-update attention examples on Maze-30$\times$30.}
    \label{fig:maze_core_examples}
\end{figure}

\Cref{fig:maze_core_example_app} shows an error-adjacent query cell (puzzle \texttt{p0040}, cycle~4, head~7, layer~0). For this query, L-update is more concentrated inside $\mathcal{N}_4(q)$, whereas H-update spreads mass over a wider region. \Cref{fig:maze_core_example_app2,fig:maze_core_example_app3} show two additional queries from different puzzles, cycles, and heads with the same qualitative pattern.

\subsection{Temporal persistence}\label{app:maze_temporal}

\begin{figure}[!htbp]
    \centering
    \includegraphics[width=\linewidth]{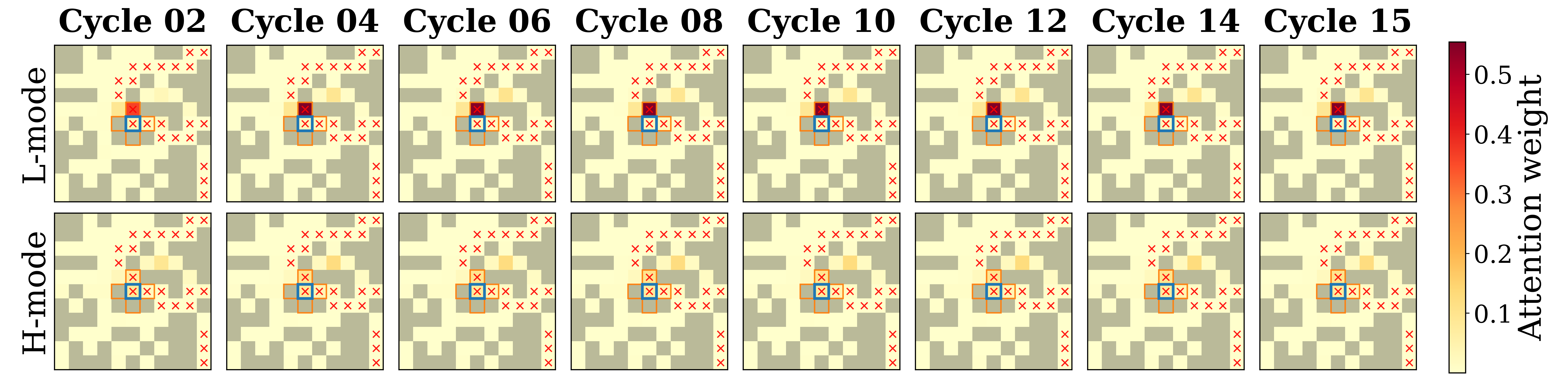}
    \caption{Temporal persistence for a single anchor query (puzzle \texttt{p0040}, head~7, layer~0, query \(r19c15\)) across cycles 2 through 15.}
    \label{fig:maze_temporal_example_app}
\end{figure}

\Cref{fig:maze_temporal_example_app} tracks one query cell, head, and layer across multiple cycles. The L-update/H-update separation persists over time: L-update repeatedly concentrates on $\mathcal{N}_4(q)$, while H-update remains broader. This rules out a one-step-artifact interpretation.

\FloatBarrier
\subsection{Layer robustness}\label{sec:maze_layers}

\Cref{fig:maze_quant} in \Cref{sec:mechanism} shows the L-update/H-update separation across all four layers; \Cref{tab:maze_layer_robustness} gives the corresponding numerical values with 95\% confidence intervals.

\begin{table}[ht]
\centering
\caption{Maze: L-update/H-update attention separation across all Transformer layers, head-averaged over the 8 heads per candidate query (puzzle-level mean $\pm$ 95\% CI; $n{=}297$ error-adjacent, $n{=}746$ control).}
\label{tab:maze_layer_robustness}
\small
\setlength{\tabcolsep}{4pt}
\begin{tabular}{llccccc}
\toprule
\textbf{Class} & \textbf{Layer} & $\Delta\rho^{(4)}$ & $\Delta\rho^{(8)}$ & $\Delta\rho^{(5\times5)}$ & $\Delta_{\mathrm{ent}}$ & $\Delta_{\mathrm{viol}}$ \\
\midrule
Err.-adj. & 0 & $0.0037_{\pm 0.0004}$ & $0.0028_{\pm 0.0003}$ & $0.0019_{\pm 0.0001}$ & $0.149_{\pm 0.005}$ & $0.006_{\pm 0.001}$ \\
          & 1 & $0.0085_{\pm 0.0007}$ & $0.0039_{\pm 0.0004}$ & $0.0044_{\pm 0.0002}$ & $0.186_{\pm 0.006}$ & $0.014_{\pm 0.002}$ \\
          & 2 & $0.0046_{\pm 0.0005}$ & $0.0047_{\pm 0.0004}$ & $0.0019_{\pm 0.0002}$ & $0.145_{\pm 0.006}$ & $0.008_{\pm 0.002}$ \\
          & 3 & $0.0412_{\pm 0.0014}$ & $0.0286_{\pm 0.0009}$ & $0.0126_{\pm 0.0004}$ & $0.314_{\pm 0.010}$ & $0.081_{\pm 0.005}$ \\
\midrule
Control   & 0 & $0.0044_{\pm 0.0002}$ & $0.0031_{\pm 0.0002}$ & $0.0017_{\pm 0.0001}$ & $0.152_{\pm 0.002}$ & $0.000_{\pm 0.000}$ \\
          & 1 & $0.0115_{\pm 0.0002}$ & $0.0057_{\pm 0.0001}$ & $0.0045_{\pm 0.0001}$ & $0.208_{\pm 0.002}$ & $0.000_{\pm 0.000}$ \\
          & 2 & $0.0026_{\pm 0.0002}$ & $0.0024_{\pm 0.0001}$ & $0.0013_{\pm 0.0001}$ & $0.157_{\pm 0.002}$ & $0.000_{\pm 0.000}$ \\
          & 3 & $0.0367_{\pm 0.0007}$ & $0.0270_{\pm 0.0005}$ & $0.0131_{\pm 0.0002}$ & $0.401_{\pm 0.003}$ & $0.000_{\pm 0.000}$ \\
\bottomrule
\end{tabular}
\end{table}

The local-versus-global split is visible throughout the network: $\Delta\rho^{(k)} > 0$ and $\Delta_{\mathrm{ent}} > 0$ at every layer for both classes, with the gap widening sharply at layer~3. The class-level $\Delta_{\mathrm{viol}}$ is positive on error-adjacent cells and grows with depth, but the within-neighborhood control reported in \Cref{sec:mechanism} is null at every layer. We therefore interpret locality on Maze as an early and robust property of L-updates, while the deeper-layer violation-specific role observed on Sudoku does not replicate.

\FloatBarrier
\subsection{Causal attention interventions}\label{app:maze_causal_attention}

\paragraph{Intervention protocol.}
We next test whether the local-versus-global attention split contributes to Maze solving, using five independently trained checkpoints and the same 1,000 test mazes for every condition.
We keep model parameters fixed and apply the knockout and diffuse interventions defined in Appendix Section \mbox{\ref{app:sudoku_causal_attention}}.
Here, the local neighborhood is $\mathcal{N}_{5\times5}(q)$: the surrounding $5\times5$ window, excluding the query itself and clipped at grid boundaries (up to 24 neighbors).
Knockout either forbids access to this neighborhood, confines access to it, or permits a random set of the same size.
Self-attention and prefix-token access remain intact in each mask; diffuse attention spreads mass uniformly over all sequence positions.
Interventions affect all heads and layers of the selected updates, with the final H-update of every recurrent cycle left unchanged to avoid directly masking the update that produces the decoded output.

\mbox{\Cref{tab:maze_causal_attention}} reports final per-cell accuracy and exact solve as mean $\pm$ sample SD across five seeds.
Changes below are in percentage points relative to this table's baseline.
We use the same halting procedure and seed-level Welch testing protocol as in the Sudoku study: one-sided tests for predicted degradation from baseline and two-sided tests otherwise.

\begin{table}[!htbp]
\centering
\caption{Maze causal attention interventions at inference (five-seed mean $\pm$ SD).}
\label{tab:maze_causal_attention}
\begin{tabular}{@{}lrr@{}}
\toprule
Intervention & Per-cell accuracy & Exact solve \\
\midrule
None (baseline) & $99.0_{\pm 0.3}$ & $71.0_{\pm 6.3}$ \\
L diffuse & $87.9_{\pm 3.8}$ & $0.1_{\pm 0.2}$ \\
L excludes neighborhood & $84.5_{\pm 2.2}$ & $0.0_{\pm 0.0}$ \\
L local only & $93.8_{\pm 1.8}$ & $1.6_{\pm 2.3}$ \\
L random only & $84.6_{\pm 1.9}$ & $0.0_{\pm 0.0}$ \\
H diffuse & $98.9_{\pm 0.3}$ & $67.3_{\pm 7.8}$ \\
H excludes neighborhood & $98.9_{\pm 0.3}$ & $68.5_{\pm 7.2}$ \\
H local only & $98.6_{\pm 0.5}$ & $55.4_{\pm 14.0}$ \\
L local, H diffuse & $92.0_{\pm 1.5}$ & $0.1_{\pm 0.2}$ \\
L diffuse, H local & $87.8_{\pm 2.8}$ & $0.0_{\pm 0.0}$ \\
Both diffuse & $84.3_{\pm 1.9}$ & $0.0_{\pm 0.0}$ \\
\bottomrule
\end{tabular}
\end{table}
\FloatBarrier

\paragraph{L's neighborhood access contributes causally to solving.}
Making L diffuse reduces per-cell accuracy by 11.1 percentage points ($p=0.001$); excluding its neighborhood reduces it by 14.5 points ($p=6\times10^{-5}$).
Both nearly eliminate exact solving (\mbox{\Cref{tab:maze_causal_attention}}).
As on Sudoku, these interventions provide causal evidence that L's neighborhood access contributes to solving.

\paragraph{Which cells L attends to matters beyond how many.}
Confining L to the true neighborhood retains higher per-cell accuracy than confining it to a random set of the same size ($93.8\%$ versus $84.6\%$, $p<10^{-4}$).
The corresponding reductions from baseline are 5.2 and 14.4 percentage points.
Both conditions expose the same number of cells per query, so the difference supports the importance of their spatial arrangement.
Nevertheless, local-only L attention achieves only $1.6\%$ exact solve: preserving neighborhood access helps cell-level predictions but is insufficient to preserve whole-maze solving.

\paragraph{H's broader attention access supports whole-maze solving.}
Confining H to its neighborhood lowers exact solve by 15.6 percentage points, from $71.0\%$ to $55.4\%$ ($p=0.033$), while changing per-cell accuracy by only $-0.4$ points ($p=0.19$).
Making H diffuse produces a smaller exact-solve decrease of 3.7 points ($67.3\%$, $p=0.43$), and excluding its neighborhood yields $68.5\%$ ($p=0.57$).
The loss in whole-maze solving under local confinement supports a contribution from H's attention beyond the neighborhood.
This is consistent with H integrating broader context to coordinate a complete solution, even when most cell predictions remain correct.

\paragraph{The assignment of attention patterns matters.}
Keeping L local and H diffuse gives higher per-cell accuracy than the swapped assignment ($92.0\%$ versus $87.8\%$, $p=0.026$).
The same two patterns therefore have different effects depending on their assignment to L and H, supporting a directional division of labor.
Exact solve is near zero for both assignments, so this comparison establishes a difference in retained per-cell accuracy rather than whole-maze solving.
\FloatBarrier

\section{Per-seed freeze results}\label{app:per_seed}\label{app:per_seed_freeze}

\begin{figure}[!htbp]
    \centering
    \includegraphics[width=\linewidth]{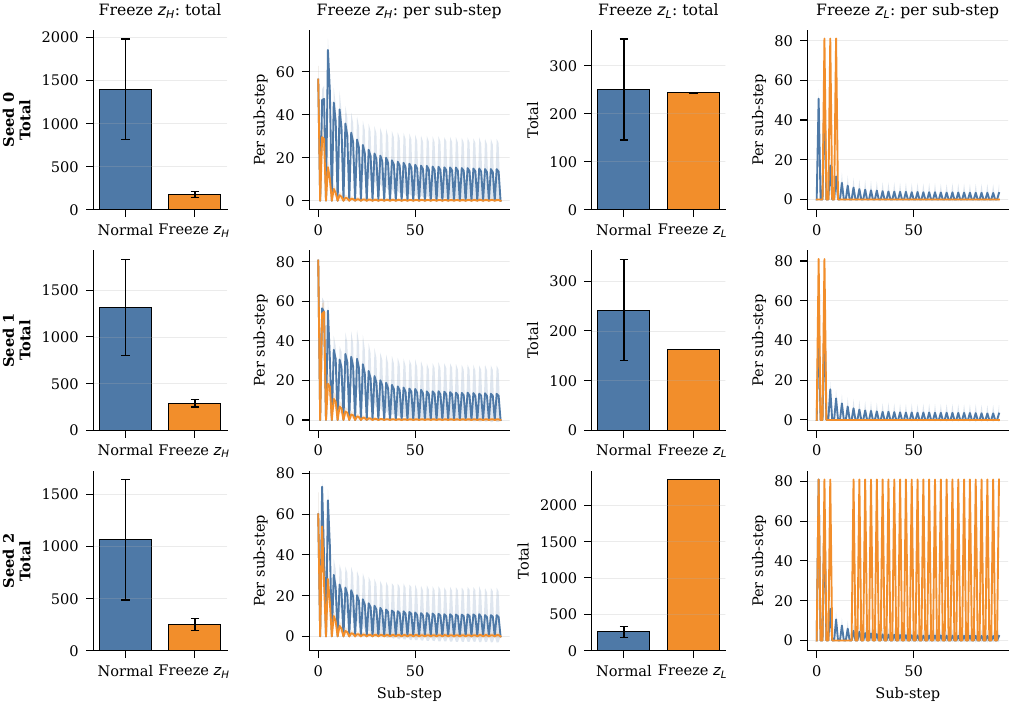}
    \caption{Per-seed freeze experiments for AIR (\texttt{Lx\_H}) on Sudoku. Each row is one seed. Bars show total content changes per puzzle and curves show per-sub-step content changes for normal (blue) and frozen (orange) rollouts; error bars and bands denote $\pm$ SD across the 1{,}000 evaluated puzzles of that seed.}
    \label{fig:per_seed_freeze_sudoku_air}
\end{figure}

\begin{figure}[!htbp]
    \centering
    \includegraphics[width=\linewidth]{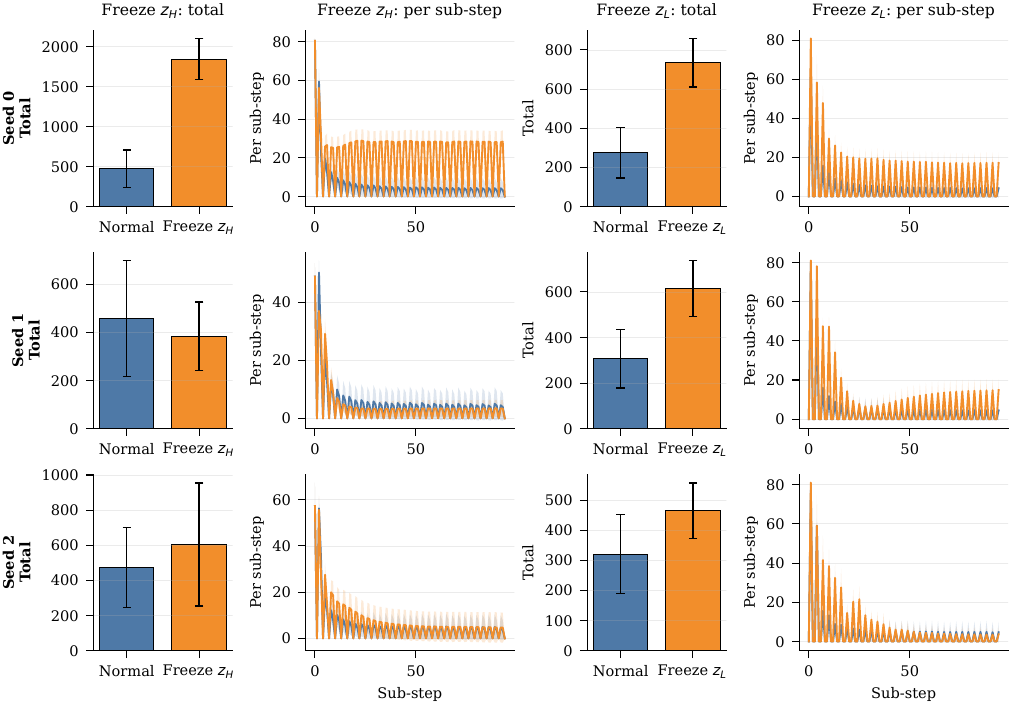}
    \caption{Per-seed freeze experiments for the symmetric \texttt{Lx\_Hx} control on Sudoku, shown as in \Cref{fig:per_seed_freeze_sudoku_air}.}
    \label{fig:per_seed_freeze_sudoku_sym}
\end{figure}

\begin{figure}[!htbp]
    \centering
    \includegraphics[width=\linewidth]{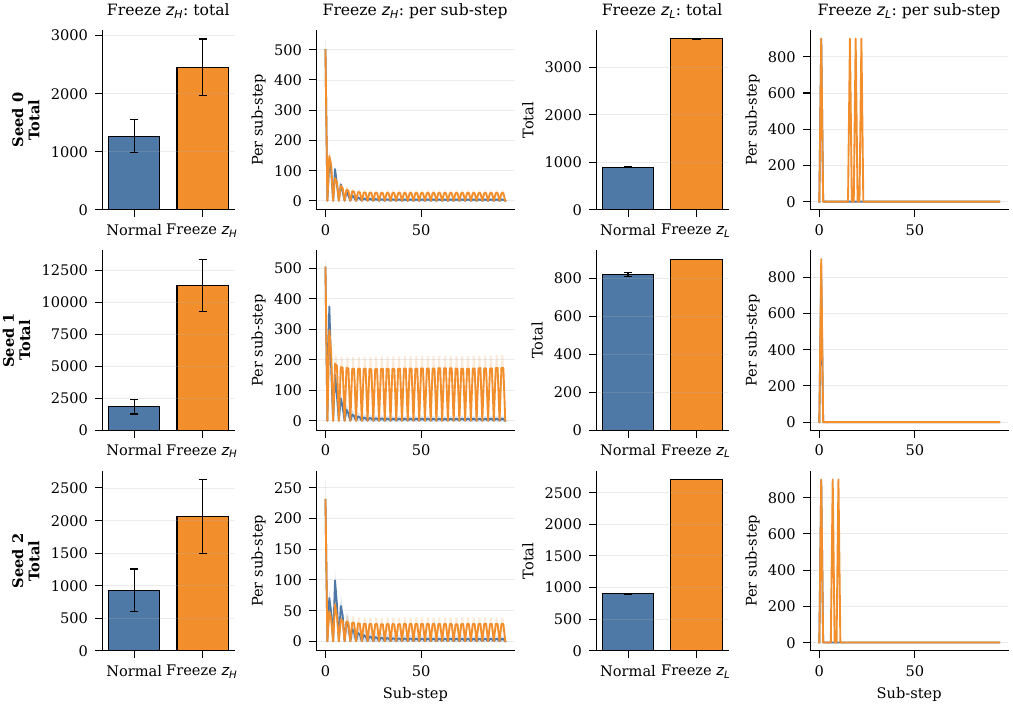}
    \caption{Per-seed freeze experiments for AIR (\texttt{Lx\_H}) on Maze, shown as in \Cref{fig:per_seed_freeze_sudoku_air}.}
    \label{fig:per_seed_freeze_maze_air}
\end{figure}

\begin{figure}[!htbp]
    \centering
    \includegraphics[width=\linewidth]{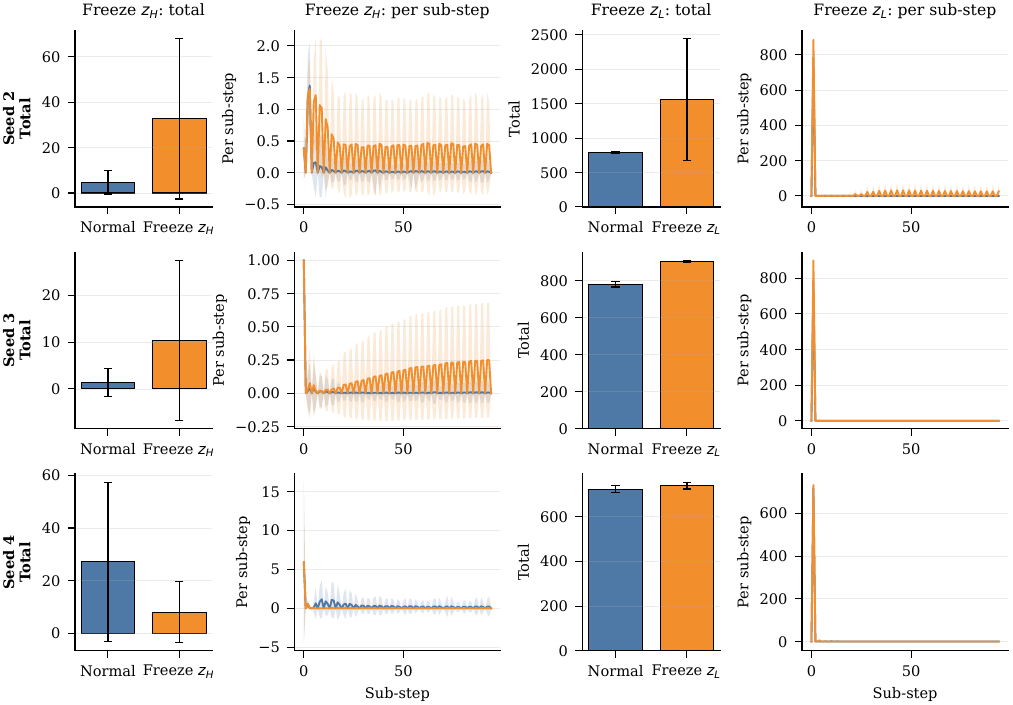}
    \caption{Per-seed freeze experiments for the symmetric \texttt{Lx\_Hx} control on Maze (seeds 2--4), shown as in \Cref{fig:per_seed_freeze_sudoku_air}.}
    \label{fig:per_seed_freeze_maze_sym}
\end{figure}

\Cref{sec:causality} reports content changes as mean $\pm$ sample SD across three seeds.
\Cref{tab:per_seed_freeze} lists each seed's mean total content change over its 1{,}000 evaluated puzzles, and \Cref{fig:per_seed_freeze_sudoku_air,fig:per_seed_freeze_sudoku_sym,fig:per_seed_freeze_maze_air,fig:per_seed_freeze_maze_sym} show the corresponding per-seed totals and per-sub-step curves.
AIR and the symmetric Sudoku control use seeds 0--2; the symmetric Maze control uses seeds 2--4.

\begin{table}[!htbp]
\centering
\caption{Per-seed freeze results: total content changes per puzzle in the unfrozen state, averaged over 1{,}000 evaluated puzzles per seed. $\Delta$ = frozen $-$ normal.}
\label{tab:per_seed_freeze}
\small
\begin{tabular}{lllrrrrrr}
\toprule
& & & \multicolumn{3}{c}{\textbf{Freeze $z_H$ (change in $z_L$)}} & \multicolumn{3}{c}{\textbf{Freeze $z_L$ (change in $z_H$)}} \\
\cmidrule(lr){4-6}\cmidrule(lr){7-9}
\textbf{Task} & \textbf{Model} & \textbf{Seed} & Normal & Frozen & $\Delta$ & Normal & Frozen & $\Delta$ \\
\midrule
Sudoku & AIR       & 0 & 1396.4 & 177.2  & $-1219.2$ & 250.3 & 243.0  & $-7.3$    \\
       &           & 1 & 1317.5 & 291.8  & $-1025.7$ & 242.1 & 162.0  & $-80.1$   \\
       &           & 2 & 1060.6 & 249.5  & $-811.1$  & 257.8 & 2349.0 & $+2091.2$ \\
\addlinespace[2pt]
       & Symmetric & 0 & 474.7  & 1847.5 & $+1372.8$ & 274.4 & 734.1  & $+459.7$  \\
       &           & 1 & 456.6  & 384.1  & $-72.5$   & 306.3 & 613.9  & $+307.6$  \\
       &           & 2 & 473.1  & 604.9  & $+131.8$  & 320.5 & 465.0  & $+144.5$  \\
\midrule
Maze   & AIR       & 0 & 1265.3 & 2453.8  & $+1188.5$ & 899.4 & 3600.0 & $+2700.6$ \\
       &           & 1 & 1840.2 & 11318.6 & $+9478.4$ & 819.5 & 900.0  & $+80.5$   \\
       &           & 2 & 931.9  & 2058.8  & $+1126.9$ & 896.7 & 2700.0 & $+1803.3$ \\
\addlinespace[2pt]
       & Symmetric & 2 & 4.8    & 32.7    & $+28.0$   & 786.9 & 1557.0 & $+770.1$  \\
       &           & 3 & 1.4    & 10.3    & $+8.9$    & 779.3 & 902.3  & $+123.0$  \\
       &           & 4 & 27.2   & 8.0     & $-19.2$   & 724.3 & 738.8  & $+14.5$   \\
\bottomrule
\end{tabular}
\end{table}

The sign of the response is consistent across seeds for five of the eight task--model--direction combinations, including freezing $z_H$ in AIR, which reduces $z_L$'s content changes on Sudoku and increases them on Maze in every seed. It is not consistent for freezing $z_L$ in AIR on Sudoku or freezing $z_H$ in the symmetric control on either task, and magnitudes vary widely throughout (\Cref{tab:per_seed_freeze}).
In the symmetric control, freezing $z_L$ increases $z_H$'s content changes in every seed on both tasks. Freezing $z_H$ increases $z_L$'s content changes in two Sudoku seeds but decreases them in the third, and changes them by fewer than 30 per puzzle in every Maze seed.
These per-seed results reinforce our conclusion in \Cref{sec:causality}: the freeze study establishes dynamical interdependence, but the direction and size of the response vary across tasks, intervention directions, and seeds.

\FloatBarrier

\section{Interpretation of task and layer differences}\label{app:interpretation}

\Cref{sec:causality,sec:mechanism} show differences between the two update types, across layers, and between tasks. We offer the following interpretation; it is consistent with our measurements but not directly tested.

\textbf{L- versus H-updates.} Only L-updates receive the encoded input $\xemb$, so L stays anchored to the given clues and the constraints around each cell. H receives no direct input and integrates information across the whole grid through the recurrent states. This matches the more local attention of L-updates and the broader attention of H-updates (\Cref{sec:mechanism}).

\textbf{Across layers.} On both tasks, L-updates are already more local and more concentrated than H-updates in the first layer, and the gap widens in deeper layers (\Cref{tab:layer_robustness} in \Cref{app:attention_layer_robustness} for Sudoku; \Cref{tab:maze_layer_robustness} in \Cref{sec:maze_layers} for Maze).
The tasks differ in what emerges with depth. On Sudoku, extra attention to violated cells appears only in the deeper layers, both in $\Delta_{\mathrm{viol}}$ and in the within-neighborhood control of \Cref{sec:mechanism}.
On Maze, $\Delta_{\mathrm{viol}}$ also grows with depth, but the within-neighborhood control stays near zero at every layer, so this growth reflects stronger locality rather than attention to the error cells themselves.
General locality therefore emerges early on both tasks, whereas attention to the specific cells in a conflict emerges in deeper layers only on Sudoku.

\textbf{Sudoku versus Maze.} The tasks differ in how their rules connect cells. A Sudoku cell must agree with every cell in its row, column, and box, so finding which of those cells conflict is useful, consistent with the violation-specific attention in deeper layers. A Maze path cell only needs to connect to its neighbors, which local attention already covers. Accordingly, L-updates attend more than H-updates to erroneous cells near the query only because they attend more locally overall; within the neighborhood, they favor erroneous cells no more than H-updates do. Many small detours also remain valid paths in Maze, so when one state is frozen the other can keep rerouting around it; this is consistent with freezing either state increasing content changes on Maze, whereas on Sudoku freezing $z_H$ suppresses $z_L$'s changes (\Cref{sec:causality}).

\end{document}